\documentclass{article}
\usepackage{hyperref}
\usepackage{url}
\usepackage{amsmath,amssymb,amsthm,mathrsfs,amsfonts,dsfont} 
\usepackage[a4paper, total={6in, 8.5in}]{geometry}
\usepackage{dsfont}
\usepackage{cite}
\usepackage{graphicx} 
\usepackage{caption}
\usepackage{subcaption}
\usepackage{algorithm}
\usepackage{algpseudocode}
\usepackage{epstopdf}
\usepackage {tikz}
\usetikzlibrary {positioning}
\definecolor {processblue}{cmyk}{0.96,0,0,0}

\newtheorem{lemma}{Lemma}

\DeclareMathOperator*{\argmax}{\arg\!\max}
\DeclareMathOperator*{\argmin}{\arg\!\min}
\DeclareMathOperator*{\sign}{sign}
\newcommand{\RNum}[1]{\uppercase\expandafter{\romannumeral #1\relax}}

\begin{document}



\title{Unsupervised Ensemble Learning with Dependent Classifiers}
\author
{Ariel Jaffe$^1 \footnote{Email addresses: Ariel Jaffe: \url{ariel.jaffe@weizmann.ac.il}, Ethan Fetaya: \url{ethan.fetaya@weizmann.ac.il},Boaz Nadler: \url{boaz.nadler@weizmann.ac.il},Tingting Jiang: \url{tingting.jiang@yale.edu},Yuval Kluger: \url{yuval.kluger@yale.edu}}$, Ethan Fetaya$^1$, Boaz Nadler$^1$,
	Tingting Jiang$^2$ and Yuval Kluger$^{2,3}$
	 \vspace{0.2cm} \\
	$^1${\footnotesize Dept. of Computer Science and Applied Mathematics, Weizmann Institute of Science, Rehovot Israel 76100}\\
	$^2${\footnotesize Interdepartmental Program in Computational Biology and Bioinformatics, Yale University, New Haven, CT 06511}\\
	$^3${\footnotesize Dept. of Pathology and Yale Cancer Center, Yale University School of Medicine, New Haven, CT 06520, USA}\\
}

\date{}
\maketitle
%

\newcommand{\fix}{\marginpar{FIX}}
\newcommand{\new}{\marginpar{NEW}}



\begin{abstract}
 In unsupervised ensemble learning, one obtains predictions from multiple sources or classifiers, yet without knowing the reliability and expertise of each source, and with no labeled data to assess it. The task is to combine these possibly conflicting predictions into an accurate meta-learner. 
Most works to date assumed perfect diversity between the different sources, a property known as conditional independence. In  realistic scenarios, however, this assumption is often violated, and ensemble learners based on it can be severely sub-optimal. The key challenges we address in this paper are:\ (i) how to detect, in an unsupervised manner, strong violations of conditional independence; and (ii) construct a suitable meta-learner. To this end we introduce a statistical model that allows for dependencies between classifiers. Our main contributions are the development of novel unsupervised methods to detect strongly dependent classifiers, better estimate  their accuracies, and construct an improved meta-learner. Using both artificial and real datasets, we showcase the importance of taking classifier dependencies into account and the competitive performance of our approach.

\end{abstract}

\section{Introduction}
In recent years unsupervised ensemble learning has become increasingly popular.
In multiple application domains one obtains the predictions, over a large set of unlabeled instances, of an ensemble of different experts or classifiers with unknown reliability. Common tasks are to
combine these possibly conflicting predictions into an accurate meta-learner, as well as assessing the accuracy of the various experts, both without any labeled data.   

A leading example is crowdsourcing, whereby a tedious labeling task is distributed to many annotators. Unsupervised ensemble learning is of increasing interest also in computational biology,  where recent works in the field propose to solve difficult prediction tasks by applying multiple algorithms and merging their results \cite{boutros2014toward,ewing2015combining,micsinai_2012,aghaeepour2013}. 
 Additional examples of unsupervised ensemble learning appear, among others, in medicine \cite{Lee_2013} and decision science \cite{quinn_2014}.


Perhaps the first to address ensemble learning in this fully unsupervised setup were Dawid and Skene \cite{Dawid_79}. A key assumption in their work was of perfect diversity between the different classifiers. Namely, their labeling errors were assumed statistically independent of each other. This property, known as \textit{conditional independence} is illustrated in the graphical model of Fig. \ref{Fig:graph} (left). In \cite{Dawid_79}, Dawid and Skene proposed to estimate the parameters of the model, i.e. the accuracies of the different classifiers, by the EM\ procedure on the non-convex likelihood function. With the increasing popularity of crowdsourcing and other unsupervised ensemble learning applications, there has been a surge of interest in this line of work, and multiple extensions of it  \cite{Welinder_2010,Karger_2011,Raykar,Whitehill_2009,Sheshadri_SQUARE_2013}. As the quality of the solution found by the EM algorithm critically depends  on its starting point, several recent works derived computationally efficient spectral methods to suggest a good initial guess \cite{Anand_2012_a,Jain_2014,Parisi_2014,Jaffe_2015}.


%

Despite its popularity and usefulness, the model of Dawid and Skene
has several limitations. One notable limitation is its assumption that 
all instances are equally difficult, with each classifier having the same probability of error over all instances. This issue was addressed, for example, by Whitehill et. al. \cite{Whitehill_2009} who introduced a model of instance difficulty, and also by  Tian et. al. 
\cite{tian2015uncovering} who proposed a model where instances are divided into groups, and the expertise of each classifier is group dependent.

A second limitation, at the focus of our work, is 
the assumption of perfect conditional independence between all classifiers. As we illustrate below, this assumption may be
strongly violated  in real-world scenarios. Furthermore, as shown in Sec. \ref{sec:exp}, neglecting classifier dependencies may yield quite sub-optimal predictions. Yet, to the best of our knowledge, relatively few works have attempted to address this important issue. 

To handle classifier dependencies, Donmez et. al. 
 \cite{Donmaz_2010} proposed a model with pairwise interactions between all classifier outputs. However, they noted that empirically, their model did not yield more accurate predictions. 
Platanios et. al. \cite{PLATANIOS_14}  developed a method to estimate the error rates of  either dependent or independent classifiers. Their method is based on analyzing the agreement rates between pairs or larger subsets of classifiers, together with a soft prior on weak dependence amongst them.

The present work is partly motivated by the ongoing somatic mutation DREAM (Dialogue for Reverse Engineering Assessments and Methods) challenge, a sequence of open competitions for detecting irregularities in the DNA string. This is a real-world example of unsupervised ensemble learning, where participants in the currently open competition are given access 
 to the predictions of more than 100 different classifiers, over more than 100,000 instances.
These classifiers were constructed by various labs worldwide, each employing their own biological knowledge and possibly proprietary labeled data. The task is to construct, in an unsupervised fashion, an accurate ensemble learner. 

In figure \ref{Fig:S1_rho} we present the empirical \emph{conditional} covariance matrix between different classifiers in one of the databases of the DREAM challenge, for which ground truth labels have been disclosed. Under the conditional independence assumption, the population conditional covariance between every two classifiers should be exactly zero. Figure  \ref{Fig:S1_rho}, in contrast, exhibits strong dependencies between groups of classifiers.  

Unsupervised ensemble learning in the presence of possibly strongly dependent classifiers raises the following two key challenges: (i) detect, in an unsupervised manner, strong violations of conditional independence; and (ii) construct a suitable meta-learner.

To cope with these challenges, in Sec. \ref{sec:setup} we introduce a new model for the joint distribution of all classifiers which allows for dependencies between them through an intermediate layer of latent variables. This generalizes the model of Dawid and Skene, and allows for groups of strongly correlated classifiers, as observed for example in the DREAM data.

In Sec. \ref{sec:detection} we devise a simple algorithm to detect subsets of strongly dependent classifiers using only their predictions and no labeled data. This is done by exploiting the structural low-rank properties of the classifiers' covariance matrix.
Figure \ref{Fig:S1_score}\ shows our resulting estimate for deviations from conditional independence on the same data as figure \ref{Fig:S1_rho}. Comparing the two figures illustrates the ability of our method to detect strong dependencies with no labeled data.  

In Sec. \ref{sec:meta} we propose methods to better estimate the accuracies of the classifiers and construct an improved meta-learner, both in the presence of strong dependencies between some of the classifiers. Finally, in Sec. \ref{sec:exp} we illustrate the competitive performance of the modified ensemble-learner derived from our model on both artificial data, four datasets from the UCI\ repository and three datasets from the DREAM\ challenge. These empirical results showcase the limitations of the strict conditional independence model, and highlight the importance of modeling the statistical dependencies between different classifiers in unsupervised ensemble learning scenarios.

\label{sec:setup}
\section{Problem Setup}
\label{sec:setup}
\paragraph{Notations.}
We consider the following binary classification problem.
Let $ \mathcal X$ be an instance space with an output space
$ \mathcal Y = \{-1, 1\}$. A labeled instance $(x, y) \in \mathcal X \times \mathcal Y$ is
a realization of the random variable $(X, Y )$. The joint distribution $p(x, y)$, as well as
the marginals $p_X(x)$ and $p_Y (y)$, are all unknown. We further
denote by $b$ the class imbalance of $Y$,
\begin{equation}
b = p_Y(1)-p_Y(-1).
\end{equation}
Let $\{f_i\}_{i=1}^m$ be a set of $m$ binary classifiers operating on $\mathcal{X}$. As our classification problem is binary, the accuracy of the $i$-th classifier is fully characterized by its sensitivity $\psi_i$ and specificity $\eta_i$,
\begin{equation}
\psi_i= \Pr\left(f_i(X)=1|Y=1\right) \qquad \eta_i= \Pr\left(f_i(X)=-1|Y=-1\right). 
\end{equation}
For future use, we denote by $\pi_{i}$ its balanced accuracy, given by the average of its sensitivity and specificity
\begin{equation}
\pi_i=\tfrac{1}{2}(\psi_i+\eta_i).
\end{equation}
Note that when the class imbalance is zero, $\pi_i$ is simply the overall accuracy of the \(i\)-th classifier.

\paragraph{The classical conditional independence model.}

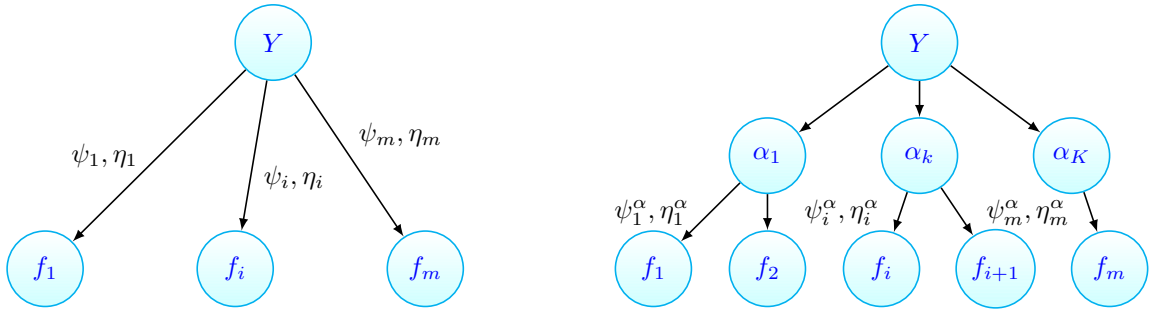
\begin{figure*}[t]
\begin{center}
\begin {tikzpicture}[-latex ,auto ,node distance =4 cm and 5cm ,on grid ,
semithick ,
state/.style ={ circle ,top color =white , bottom color = processblue!20 ,
        draw,processblue , text=blue , minimum width =1 cm}]
\node[state] (Y) at (2,3) {$Y$};
\node[state] (f1) at (-1,0) {$f_1$};
\node[state] (fi) at (1.5,0) {$f_i$};
\node[state] (fm) at (4,0) {$f_m$};
\path (Y) edge node [above =0.15 cm,left = 0.15cm] {$\psi_1,\eta_1$} (f1);
\path (Y) edge node {$\psi_i,\eta_i$} (fi);
\path (Y) edge node {$\psi_m,\eta_m$} (fm);
\node[state] (Y2) at (10.5,3) {$Y$};
\node[state] (a1) at (8.5,1.5) {$\alpha_1$};
\node[state] (ak) at (10.5,1.5) {$\alpha_k$};
\node[state] (aK) at (12.5,1.5) {$\alpha_K$};
\node[state] (F1) at (7,0) {$f_1$};
\node[state] (F2) at (8.5,0) {$f_2$};
\node[state] (Fi) at (10,0) {$f_i$};
\node[state] (Fi2) at (11.5,0) {$f_{i+1}$};
\node[state] (Fm) at (13,0) {$f_{m}$};
\path (Y2) edge node [above =0.15 cm,left = 0.15cm] {}(a1);
\path (Y2) edge node [above =0.15 cm,left = 0.15cm] {}(ak);
\path (Y2) edge node [above =0.15 cm,left = 0.15cm] {}(aK);
\path (a1) edge node [above =0.15 cm,left = 0.15cm] {$\psi_1^\alpha,\eta_1^\alpha$}(F1);
\path (a1) edge node [above =0.15 cm,left = 0.15cm] {}(F2);
\path (ak) edge node [above =0.15 cm,left = 0.15cm] {$\psi_i^\alpha,\eta_i^\alpha$}(Fi);
\path (ak) edge node [above =0.15 cm,left = 0.15cm] {}(Fi2);
\path (aK) edge node [above =0.15 cm,left = 0.15cm] {$\psi_m^\alpha,\eta_m^\alpha$}(Fm);
\end{tikzpicture}
\caption{(Left) The perfect conditional independence model of Dawid and Skene. All classifiers are independent given the class label $Y$; (Right) The generalized model considered in this work.}
\label{Fig:graph}
\end{center}
\end{figure*}



In the model proposed by Dawid and Skene \cite{Dawid_79}, depicted in Fig. \ref{Fig:graph}(left), all $m$ classifiers were assumed conditionally independent given the class label. Namely, for any set of predictions \(a_{1},\ldots,a_m\in\{\pm1\}\)
\begin{equation}
\Pr(f_1=a_1,\ldots,f_m=a_m|Y)=\prod_i \Pr(f_i=a_i|Y)
\label{eq:ind_y}.
\end{equation}
As shown in  \cite{Dawid_79}, the maximum likelihood estimation (MLE) for $y$ given the parameters $\psi_i,\eta_i$ and $b$ is \emph{linear} in the predictions of  $f_1,...,f_m$
\begin{equation}\label{eq:DS_pred}
\hat{y} = \sign\left(\sum_{i=1}^mw_if_i(x)+w_0\right),\,\, w_i=w(\psi_i,\eta_i).
\end{equation}
Hence, the main challenge is to estimate the model parameters $\psi_i$ and $\eta_i$. A simple approach to do so, as described in \cite{Parisi_2014,Jaffe_2015}, is based on the following insight: A classifier which is totally random has zero correlation with any other classifier. In contrast, a high correlation between the predictions of two classifiers is a strong indication 
that both are highly accurate,  assuming they are not both adversarial.  

 In many realistic scenarios, however, an ensemble may contain several strongly dependent classifiers. Such a scenario has several consequences: First, the above insight that high correlation between two classifiers implies that both are accurate breaks down completely. Second, as shown in Sec. \ref{sec:exp}, estimating the classifiers parameters $\psi_i,\eta_i$ as if they were conditionally independent may be highly inaccurate. Third, in contrast to 
 Eq. (\ref{eq:DS_pred}),\ the optimal ensemble learner is in general \emph{non-linear} in the $m$ classifiers. Applying the linear meta-classifier of Eq. (\ref{eq:DS_pred}) may be suboptimal, even when provided with the true classifier accuracies.
 
 
\paragraph{A\ model for conditionally dependent classifiers.}

In this paper we significantly relax the conditional independence assumption. We introduce a new model which allows classifiers to be dependent through unobserved latent variables, and develop novel methods to learn the model parameters and construct an improved non-linear meta-learner.

In contrast to the 2-layer model of Dawid and Skene, our proposed model, illustrated in Fig. \ref{Fig:graph}(right), has an additional intermediate layer with $K\leq m$ latent binary random variables  $\{\alpha_k\}_{k=1}^K$. In this model, the unobserved $\alpha_k$ are conditionally independent given the true label $Y$, whereas each observed classifier depends on $Y$ only through a single and unknown latent variable. Classifiers that depend on different latent variables are thus conditionally independent given $Y$, whereas classifiers that depend on the same latent variable may have strongly correlated prediction errors. Each hidden variable can thus be interpreted as a separate unobserved teacher, or source of information, and the classifiers that depend on it are different perturbations of it.
Namely, even though we observe \(m\) predictions for each instance, they are in fact generated by a hidden model with intrinsic dimensionality \(K\), where possibly \(K\ll m\).

Let us now describe in detail our probabilistic model. First, since the latent variables $\alpha_1,\ldots,\alpha_K$ follow the classical model of Dawid and Skene, their joint distribution is fully characterized by the class imbalance $b$ and the $2K$ probabilities
\[
\Pr(\alpha_k=1|Y=1)\quad \mbox{and}\quad \Pr(\alpha_k=-1|Y=-1).
\]
Next,  we introduce an assignment function $\mathbf{ c}:[m]\rightarrow [K]$, such that if classifier $f_i$ depends on $\alpha_k$ then $\mathbf c(i) = k$.
The dependence of classifier $f_i$ on the class label $Y$ is \textit{only} through its latent variable $\alpha_{\mathbf c(i)}$,   
\begin{equation} \label{eq:alpha_y}
\Pr(f_i|\alpha_{\mathbf c(i)},Y)=\Pr(f_i|\alpha_{\mathbf c(i)}). 
\end{equation}

Hence, classifiers $f_i,f_{j}$ with $\mathbf{ c}(i)\neq \mathbf{ c}(j)$  maintain the original conditional independence assumption of Eq. \eqref{eq:ind_y}.
In contrast, classifiers $f_{i},f_{j}$ with $\mathbf{ c}(i)= \mathbf{ c}(j)$ are only conditionally independent given $\alpha_{\mathbf c(i)}$,
\begin{equation}
\Pr(f_i = a_i,f_{j}=a_{j}|\alpha_{\mathbf c(i)}) =  
\Pr(f_i = a_i|\alpha_{\mathbf c(i)})\Pr(f_{j}=a_{j}|\alpha_{\mathbf c(i)}). 
\label{eq:ind_alpha}
\end{equation}

Note that if the number of groups $K$ is equal to the number of classifiers, then all classifiers are conditionally independent, and we recover the original model of Dawid and Skene. 

Since the model now consists of three layers, the remaining parameters to describe it are the sensitivity $\psi_i^\alpha$ and specificity $\eta_i^\alpha$ of the \(i\)-th classifier given its latent variable $\alpha_{\mathbf c(i)}$,
\begin{equation*}
\psi_i^\alpha\! =\! \Pr(f_i=1|\alpha_{\mathbf c(i)}=1),\ \ 
\eta_i^\alpha \!=\! \Pr(f_i=\!-1|\alpha_{\mathbf c(i)}=\!-1).
\end{equation*}
By Eq. \eqref{eq:alpha_y}, the overall sensitivity $\psi_i$  of the \(i\)-th classifier is related to $\psi_i^\alpha$ and $\eta_i^\alpha$ via 
\begin{equation}\label{eq:bla}
\psi_i = \Pr(\alpha_{\mathbf c(i)}=1|Y=1)\psi_i^\alpha+
  \Pr(\alpha_{\mathbf c(i)}=-1|Y=1)(1-\eta_i^\alpha), \end{equation} 
with a similar expression for its overall specificity $\eta_i$.

\paragraph{Remark on Model Identifiability.} Note that the model depicted in Fig. \ref{Fig:graph}(right) is in general not identifiable. For example, the classical model of Dawid and Skene can also be recovered with a single latent variable $K=1$, by having $\alpha_1=Y$. Similarly, for a latent variable that has only a single classifier dependent on it, the parameters $\psi_i,\eta_i$ and $\psi^\alpha,\eta^\alpha$ are non-identifiable. Nonetheless, these non-identifiability issues do not affect our algorithms, described below. 

\paragraph{Problem Formulation.} We consider the following totally unsupervised scenario. Let $Z$ be a binary $m \times n$ matrix with entries $Z_{ij} = f_i(x_j),
$ where $f_i(x_j)$ is the label predicted by classifier $f_i$ at instance $x_j$.
We assume $x_j$ are drawn i.i.d. from $p_X(x)$. 
 We also assume the \(m\) classifiers satisfy our generalized model, but otherwise we have no prior knowledge as to the number of groups $K$, the assignment function $\mathbf c$ or the classifier accuracies (sensitivities $\psi_i$,$\psi_i^\alpha$ and 
specificities $\eta_i, \eta_i^\alpha$).
Given only the matrix $Z$ of binary predictions and no labeled data, we consider the following problems:
\begin{enumerate}
\item Is it possible to detect  strongly dependent classifiers, and estimate the number of groups and the corresponding assignment function $\mathbf c$?
\item Given a positive answer to the previous question, how can we estimate the sensitivities and specificities of the $m$ different classifiers and construct an improved, possibly non-linear, �meta learner� ?
\end{enumerate}

\section{Estimating the assignment function}
\label{sec:detection}

The main challenge in our model is the first problem of estimating the number of groups $K$ and the assignment function $\mathbf c$. Once $\mathbf c$ is obtained, we will see in Section \ref{sec:meta} that our second problem can be reduced to the conditional independent case,  already addressed in previous works \cite{Jaffe_2015,Parisi_2014,Zhang_2014,Jain_2014}. In principle, one could try to fit the whole model by maximum likelihood, however this results in a hard combinatorial problem. We propose instead to first estimate only $K$ and $\mathbf c$. We do so using the 
low-rank structure of the covariance matrix of the classifiers, implied by our model. 
\paragraph{The covariance matrix.}
\label{sub:structure}
 Let $R$ denote the $m\times m$ population covariance matrix of the $m$ classifiers
 \begin{equation}
 r_{ij} = \mathds{E}[(f_i-\mathds{E}[f_i])(f_j-\mathds{E}[f_j])].
 \label{eq:cov}
 \end{equation}

The following lemma describes its structure. It generalizes a similar lemma, for the standard Dawid and Skene model, proven in  \cite{Parisi_2014}. The proof of this and other lemmas below appear in the appendix.
\begin{lemma}
\label{lem:R}
There exists two vectors $v^{on},v^{off}\in \mathds{R}^m$ such that for all $i\neq j$,
\begin{equation}
r_{ij} = \left\{
\begin{array}{lr}
v^{off}_i\cdot v^{off}_j & \textnormal{if } \mathbf c(i)\neq\mathbf c(j)\\
v^{on}_i\cdot v^{on}_j &  \textnormal{if  } \mathbf c(i)=\mathbf c(j)
\end{array}
\right. 
\end{equation}
\end{lemma}
The population covariance matrix is therefore a combination of  two rank-one matrices.  
The block diagonal elements $i,j$ with $\mathbf c(i)=\mathbf c(j)$ correspond to the rank-one matrix $ v^{on} (v^{on})^T$, where on stands for on-block, while the off-block diagonal elements, with $\mathbf c(i)\neq\mathbf c(j)$ correspond to another rank-one matrix $ v^{off} ( v^{off})^T$.
Let us define the indicator $\mathds{1}_{\mathbf c}(i,j)$
\begin{equation}
\mathds{1}_{\mathbf c}(i,j) = 
\begin{cases}
1 & \mathbf c(i)=\mathbf c(j) \\
0 & \textnormal{otherwise}
\end{cases}
\end{equation}
The non-diagonal elements of $R$ can thus  be written as follows,
\begin{equation}
r_{ij} = \mathds{1}_{\mathbf c}(i,j)v_i^{on} v_j^{on}+(1-\mathds{1}_{\mathbf c}(i,j))v_i^{off}v_j^{off}.
\label{eq:r_ij}
\end{equation}


\paragraph{Learning the model in the  ideal setting. }
\label{sub:perfect}
It is instructive to first examine the case where the data is generated according to our model, and the population covariance matrix \(R\) is exactly known, i.e. $n=\infty$. The question of interest is whether it is possible to recover the assignment function in this setting. 

To this end, let us look at the possible values of the determinant of  2x2 submatrices of \(R,\)
\begin{equation}
M_{ijkl}=\det\left(
\begin{array}{cc}
r_{ij} & r_{il} \\
r_{kj} & r_{kl} 
\end{array}
\right)
\end{equation}
Due to the low rank structure described in lemma \ref{lem:R}, we have the following result, with the exact conditions appearing in the appendix.

\begin{lemma}\label{lem:sparsity}
Assume the two vectors $v^{on}$ and $v^{off}$ are sufficiently different, then \(M_{ijkl}=0\) if and only if either:
(i) Three or more of the indices $i,j,k$ and $l$ belong to the same group or (ii) $\mathbf c(i)\neq \mathbf c(j)$,  $\mathbf c(j)\neq \mathbf c(k)$, $\mathbf c(k)\neq \mathbf c(l)$ and $ \mathbf c(l)\neq \mathbf c(i)$. 
\end{lemma}
With details in the appendix, comparing the indices \((j,k,l)\) where $M(i_1,j,k,l)=0$ with $i_1$ fixed, to those where $M(i_2,j,k,l)=0$, we can deduce, in polynomial time, whether $\mathbf c(i_1)=\mathbf c(i_2)$.


 

\paragraph{Learning the model in practice.}
\label{sub:real}
In practical scenarios,  the population covariance matrix $R$ is unknown and we  can only
compute the sample covariance matrix $\hat{R}$. Furthermore, our model would typically be only an approximation of the classifiers dependency structure. Given only $\hat{R}$, the approach to recover the assignment function described above, based on exact matching of the pattern of zeros of the determinants of various 2x2 submatrices is clearly not applicable.

In principle, since $\mathds{E}[\hat{R}]=R$ a standard approach would be to define the following residual
\begin{equation}
\Delta(v^{on}, v^{off},\mathbf c) = \sum_{i \neq j} \mathds{1}_{\mathbf c}(i,j)(v_i^{on} v_j^{on}-\hat{r}_{ij})^2 + 
        (1-\mathds{1}_{\mathbf c}(i,j))(v_i^{off}v_j^{off} -\hat{r}_{ij})^2,
        \label{eq:res}
\end{equation}
and find its global minimum. Unfortunately, as stated in the following lemma and proven in the appendix, in  general  this is not a simple task.
\begin{lemma}
Minimizing the residual of Eq. \eqref{eq:res} for a general covariance matrix $\hat{R}$ is NP-hard. 
\label{lem:NP}
\end{lemma}
 In light of Lemma \ref{lem:NP}, we now present a tractable algorithm to estimate $K$ and $\mathbf{c}$ and provide some theoretical support for it.  Our algorithm is inspired by the ideal setting which highlighted the importance of the determinants of $2\times 2$ submatrices. 
 To detect pairs of classifiers $f_i,f_j$ that strongly violate the conditional independence assumption, we thus compute the following score matrix $\hat{S}=\hat{S}(\hat{R})$,
\begin{equation}
\hat{s}_{ij} = \sum_{k,l \neq i,j} |\hat{r}_{ij}\hat{r}_{kl}-\hat{r}_{il}\hat{r}_{kj}|.
\label{eq:score}
\end{equation}
The idea behind the score matrix is the following: Consider the score matrix $S$ computed with the population covariance $R$. Lemma \ref{lem:sparsity} characterized the cases where the submatrices in Eq. (\ref{eq:score})\ are of rank-one, and hence their determinant is zero.  When  $\mathbf c(i)\neq\mathbf c(j)$ most submatrices
come from four different groups, i.e. will have rank one, and thus the sum $s_{ij}$ will be small. On the other hand, when $\mathbf c(i)=\mathbf c(j)$ many submatrices
will not be rank one and thus $s_{ij}$ will be large, assuming no degeneracy between $ v^{on}$ and $ v^{off}$. As $\hat{S}\xrightarrow{n\rightarrow \infty} S$, large values of $\hat{s}_{ij}$ serve as an indication of strong conditional dependence between classifiers \(f_{i}\) and \(f_j\).

The following lemma provides some theoretical justification for the utility of the score matrix $S$ computed with the population covariance, in recovering the assignment function $\mathbf c$. For simplicity, we analyze the 'symmetric' case where the class imbalance $b=0$, $\Pr(\alpha_k = -1|y=-1)=\Pr(\alpha_k=1|y=1)$ and all groups have equal size of $m/K$. 
We measure deviation from conditional independence  by the following matrices of conditional covariances $C^+$ and $C^-$,
\begin{eqnarray} \label{eq:con_cor}
c^+_{ij} &=& \mathds{E}[(f_i-\mathds E[f_i])(f_j-\mathds E[f_j])|Y=1]\nonumber \\ c^-_{ij} &=&\mathds{E}[(f_i-\mathds E[f_i])(f_j-\mathds E[f_j])|Y=-1].
\end{eqnarray}
Finally, we assume there is a $\delta>0$ such that the balanced accuracies of all classifiers satisfy $(2\pi_i-1)>\delta>0$. 
\begin{lemma}
Under the assumptions described above,  if $\mathbf c(i)=\mathbf c(j)$ then 
\begin{equation}
s_{ij} > m^2\left(1-\frac{3}{K}\right)\delta^2|c^+_{ij}|=m^2\left(1-\frac{3}{K}\right)\delta^2|c^-_{ij}|.
\label{eq:bound}
\end{equation}
In contrast, if $\mathbf c(i)\neq \mathbf c(j)$ then
\begin{equation}
s_{ij} < \frac{2m^2}{K}\left(1-\frac{2}{K}\right).
\label{eq:upperBound}
\end{equation}
 \label{lem:score}
\end{lemma}
An immediate corollary from lemma \ref{lem:score}, is that if the classifiers are sufficiently accurate, and their dependencies within each group are strong enough then the score matrix exhibits a clear gap with $\max\limits_{\mathbf c(i)\neq \mathbf c(j)}S_{ij}<\min\limits_{\mathbf c(i)= \mathbf c(j)}S_{ij}$. In this case, even a simple single-linkage hierarchical clustering algorithm can recover the correct assignment function from $S$. In practice, as only $\hat{S}$ is available, we apply spectral clustering  which is more robust, and works better in practice. 



\begin{figure}[t]
\centering
        \begin{subfigure}[b]{.45\textwidth}
                \centering      
                \includegraphics[width=\textwidth]{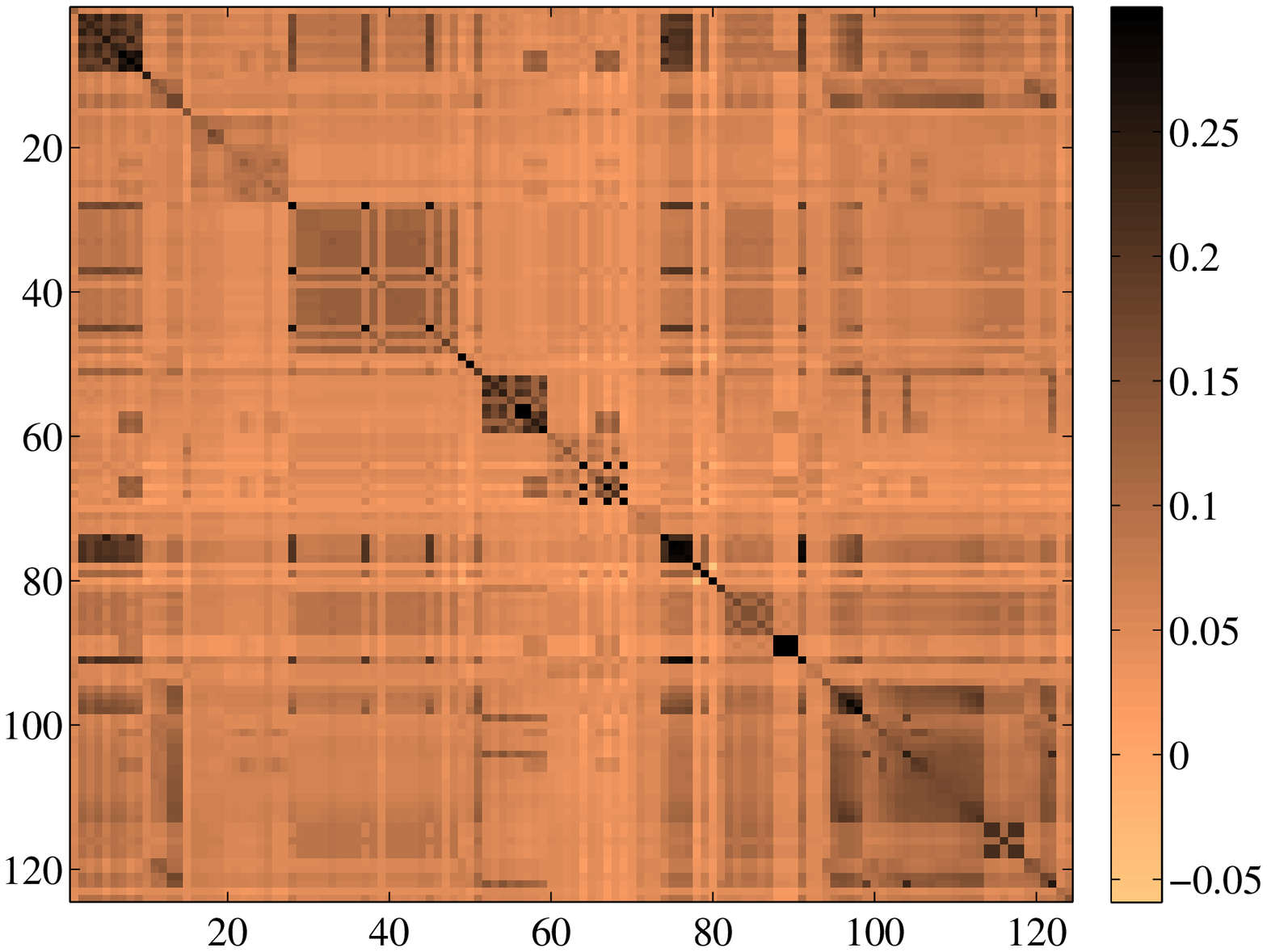}
                \caption{The \textit{conditional} covariance matrix $\tfrac{1}{2}(C^++C^-)$ of the DREAM dataset S1, computed using the ground truth labels. \\\\ }
                \label{Fig:S1_rho}
        \end{subfigure}
        \hspace{1cm}        
        \begin{subfigure}[b]{.45\textwidth}
                \centering
                \includegraphics[width=\textwidth]{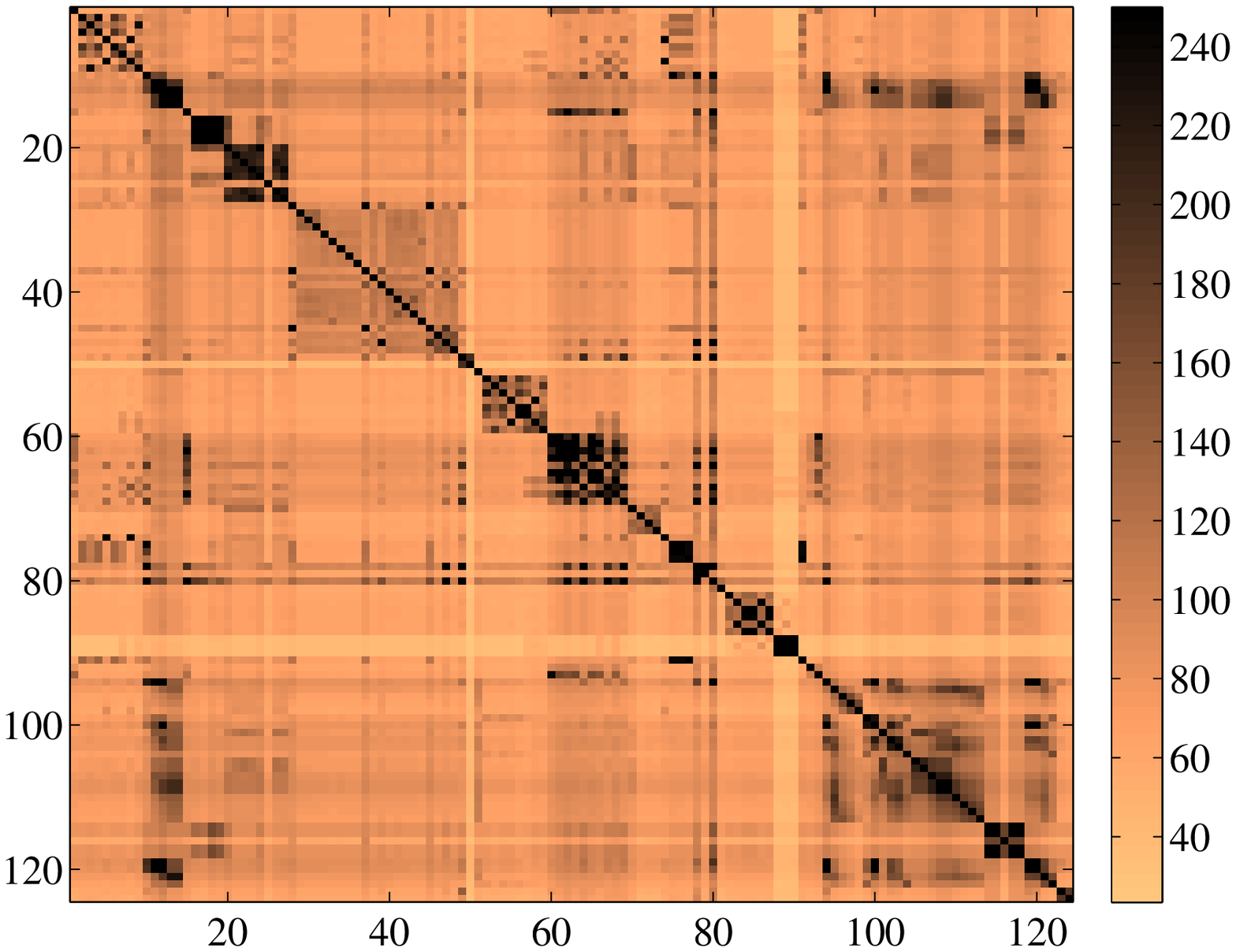}
                \caption{The score matrix $\hat{S}$ of the DREAM S1 dataset, computed from the matrix of classifier predictions. For visualization purposes, the upper limit of the above score matrix is fixed at 300.}
                \label{Fig:S1_score}
        \end{subfigure}
        \caption{}
\end{figure}

\begin{figure}[!t]        
                \centering   
                \includegraphics[width=0.5\textwidth]{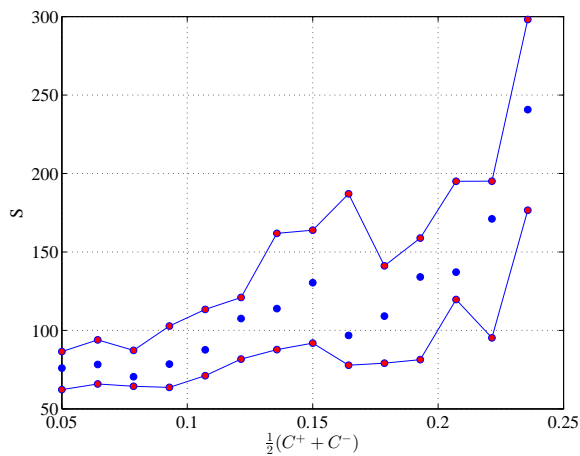}
                \caption{Values of \(S\) vs. the corresponding conditional covariance matrix $\frac{1}{2}(C^++C^-)$ for the DREAM dataset S1. The blue dots represent the mean value,  the upper and lower red dots represent the $20th$ and $80th$ quantiles, respectively.}
                \label{Fig:S_vs_a}
\end{figure}

We illustrate the usefulness of the score matrix using the DREAM challenge S1 dataset, which contains $m=124$ classifiers. Fig. \ref{Fig:S1_rho} shows the matrix of conditional covariance $\tfrac{1}{2}(C^++C^-)$ of Eq. (\ref{eq:con_cor}), computed using the ground truth labels. Fig. \ref{Fig:S1_score} shows the score matrix $\hat{S}$ computed using only the classifiers predictions. We also plot the values of the score matrix vs. the conditional covariance in figure \ref{Fig:S_vs_a}. Clearly, a high score is a reliable indication for strong conditional dependencies between classifiers.


\begin{algorithm}[t]
        \caption{Estimating the assignment function $\mathbf c$ and vectors $v^{on},v^{off}$}
        \begin{algorithmic}[1]
                \State Estimate the covariance matrix $R$ \eqref{eq:cov}.
                \State Obtain the score matrix by \eqref{eq:score}
                \ForAll{$1 < k < m$}
                        \State Estimate $\mathbf c$ by performing spectral clustering with the Laplacian of the score matrix.
                        \State Use the clustering function to estimate the two vectors $ v^{on}, v^{off}$.
                        \State Calculate residual by \eqref{eq:res}.
                \EndFor         
                \State Pick the assignment function and vectors which yield minimal residual.
        \end{algorithmic}
        \label{algo_2}
\end{algorithm}

 It is important to note that the time complexity needed to build the score matrix $S$ is $\mathcal{O}(m^4)$. While quartic scaling is usually considered too expensive, in our case as the number $m$ of classifiers in many real world problems is in the hundreds our algorithm can run on these datasets in less than an hour. This can be sped-up, for example, by sampling the elements of $S$ instead of computing the full matrix \cite{Fetaya_2015}. 

\paragraph{Estimating the assignment function $\mathbf c$.}

We estimate $\mathbf c$ by spectral clustering the score matrix $\hat{S}$ of  Eq. (\ref{eq:score}). As the number of clusters or groups $K$ is unknown, we choose the one which minimizes the residual function defined in Eq. \eqref{eq:res}.
The steps for estimating the number of groups $K$ and the assignment function $\mathbf c$ are summarized in Algorithm \ref{algo_2}.
Note that retrieving $v^{on}$ and $ v^{off}$ from the covariance matrix is a rank-one matrix completion problem, for which several solutions exist, for example see \cite{Candes_2009}.
Also note that while we compute spectral clustering for various number of clusters, the costly eigen-decoposition step only needs to be done once.

\section{The latent spectral meta learner}\label{sec:meta}

\paragraph{Estimating the model parameters.}
Given estimates of \(K\) and of the assignment function $\mathbf c$, estimating the remaining model parameters can be divided into two stages: (i) Estimating the sensitivity and specificity of the different classifiers given the latent variables $\alpha_k$:  $\psi_i^\alpha,\eta_i^\alpha$ (ii) Estimating the probabilities associated with the latent variables, $\Pr(\alpha_k=1|Y=1)$
and $\Pr(\alpha_k=-1|Y=-1)$.

The key observation is that in each of these stages the underlying model
follows the classical conditional independent model of \cite{Dawid_79}. 
In particular, classifiers with a common latent variable are conditionally independent given its value. Similarly the \(K\) latent variables themselves are conditionally independent given the true label $Y$. Thus, we can solve the two stages sequentially by any of the various methods already developed for the Dawid and Skene model. In our implementation, we used the spectral meta learner proposed in \cite{Jaffe_2015}, whose code is publicly available
. A pseudo-code for this process appears in Algorithm \ref{algo_params}.

\paragraph{Label Predictions.}

Once all the parameters of the model are known, for each instance \(x\) we estimate its label by maximum likelihood 
\begin{equation}
\hat y = \argmax_{y = \pm 1} \Pr(f_1(x), \ldots, f_m(x)|y).
\end{equation}
Following our generative model, Fig. \ref{Fig:graph}(right), the above probability is a function of the model parameters $b,\psi_i^\alpha,\eta_i^\alpha,\psi_\alpha,\eta_\alpha$, and the assignment function $\mathbf c$.

\paragraph{Classifier selection.}
In some cases, it is required to construct a sparse ensemble learner which uses only a small subset of at most $M$ out of the available \(m\) classifiers. This problem of selecting a small subset of classifiers, known as \textit{ensemble pruning}, has mostly been studied in supervised settings, see \cite{Rokach2009,martinez2009,Yin_2014}.


Under the conditional independence assumption, the best subset simply consists of the $M$ most accurate  classifiers. In our model, in contrast, the correlations between the classifiers have to be taken into account.  Assuming the required number of classifiers is smaller than the number of groups $M \leq K$, a simple approach is to select the $M$ most accurate classifiers under the constraint that they all come from different groups. This creates a balance between accuracy and diversity.

%
%
    
\begin{algorithm}[t]
\caption{Estimate model parameters}
\begin{algorithmic}[1]
\State {\bf{Input:}} Matrix of predictions $f_{i}(x_j)$, parameters \(K\) and $\mathbf{c}$.
\For{$ k = 1,..,K$}
        \State Find all classifiers $f_i$ where $\mathbf c(i)=k$
        \State Estimate $\psi_i^\alpha, \eta_i^\alpha$ and $\mathbb E[\alpha_k]$
        \State Estimate the latent values $\alpha_{k}(x_j)$, $\forall j=1,...,n$       
\EndFor
\State Estimate $\Pr(\alpha_k=1|Y=1), \Pr(\alpha_k=-1|Y=-1)$
\end{algorithmic}
\label{algo_params}
\end{algorithm}

\section{Experiments}
\label{sec:exp}

We demonstrate the performance of the latent variable model on artificial data, on datasets from the UCI repository and on the ICGA-TCGA dream challenge. 

Throughout our experiments, we compare the performance of the following unsupervised ensemble methods: (1) \texttt{Majority voting}, which serves as a baseline; (2) \texttt{SML+EM} - a spectral meta-learner based on the independence assumption \cite{Jaffe_2015} which provides an initial guess, followed by EM iterations; (3) \texttt{Oracle-CI}: 
A linear meta-learner based on Eq. (\ref{eq:DS_pred}), which assumes conditional independence but is given the exact accuracies of all the individual classifiers. (4) \texttt{L-SML} (latent SML), the new algorithm presented in this work.
 
For the artificial data, we also present the performance of its oracle meta-learner, denoted \texttt{Oracle-L}, which is given the exact structure and parameters of the model, and predicts the  label \(Y\) by maximum likelihood.


\subsection{Artificial Data}\label{sec:artifical}
\label{subsec:artificail}
\begin{figure}[!t]      
        \begin{minipage}[b]{.45\textwidth}
                \centering      
                \includegraphics[width=\textwidth]{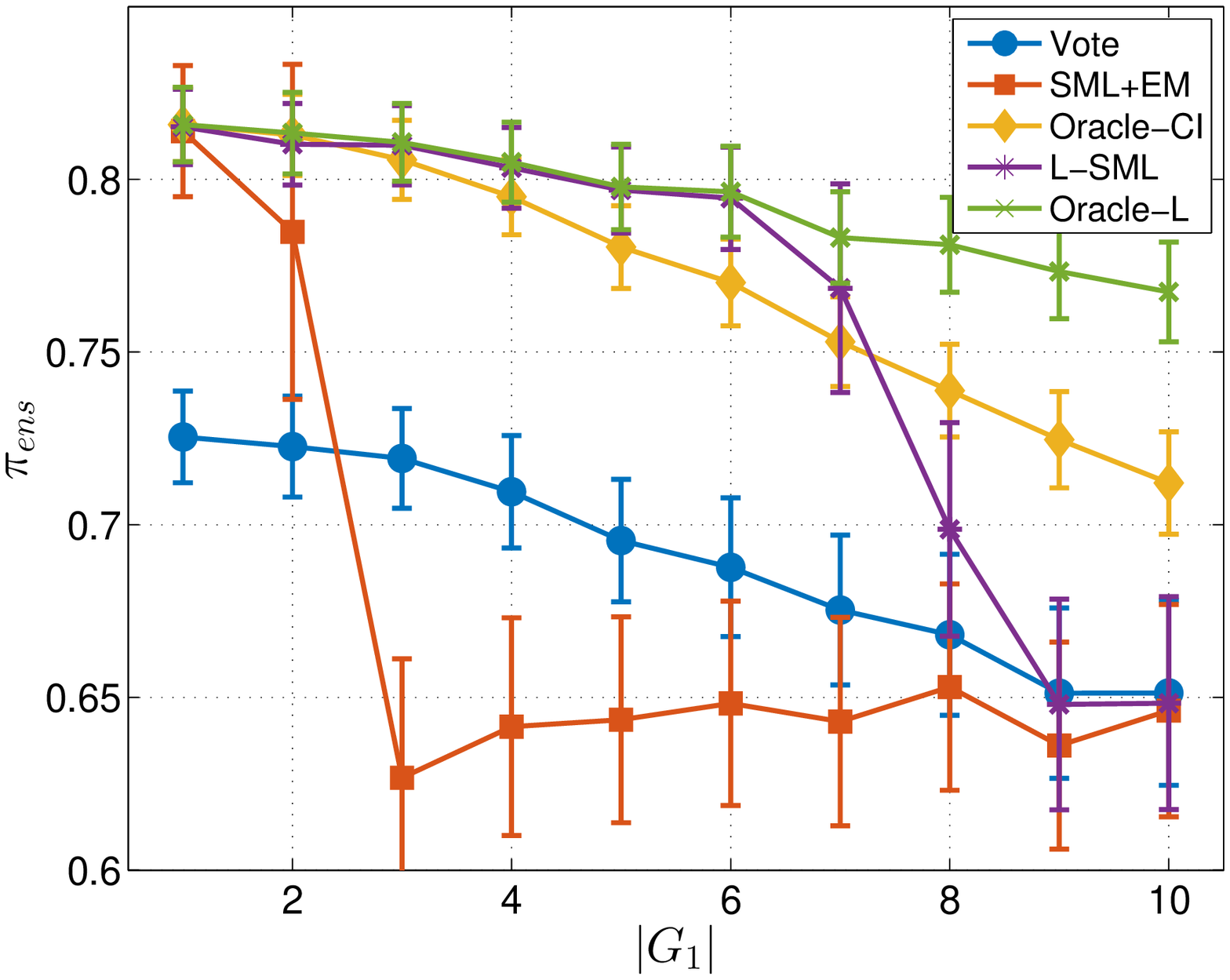}
                \caption{Simulated data: Ensemble learner balanced accuracy vs. the size of group 1.}
                \label{Fig:Simulated}
        \end{minipage}
        \hspace{1cm}
        \begin{minipage}[b]{.45\textwidth}
                \centering
                \includegraphics[width=\textwidth]{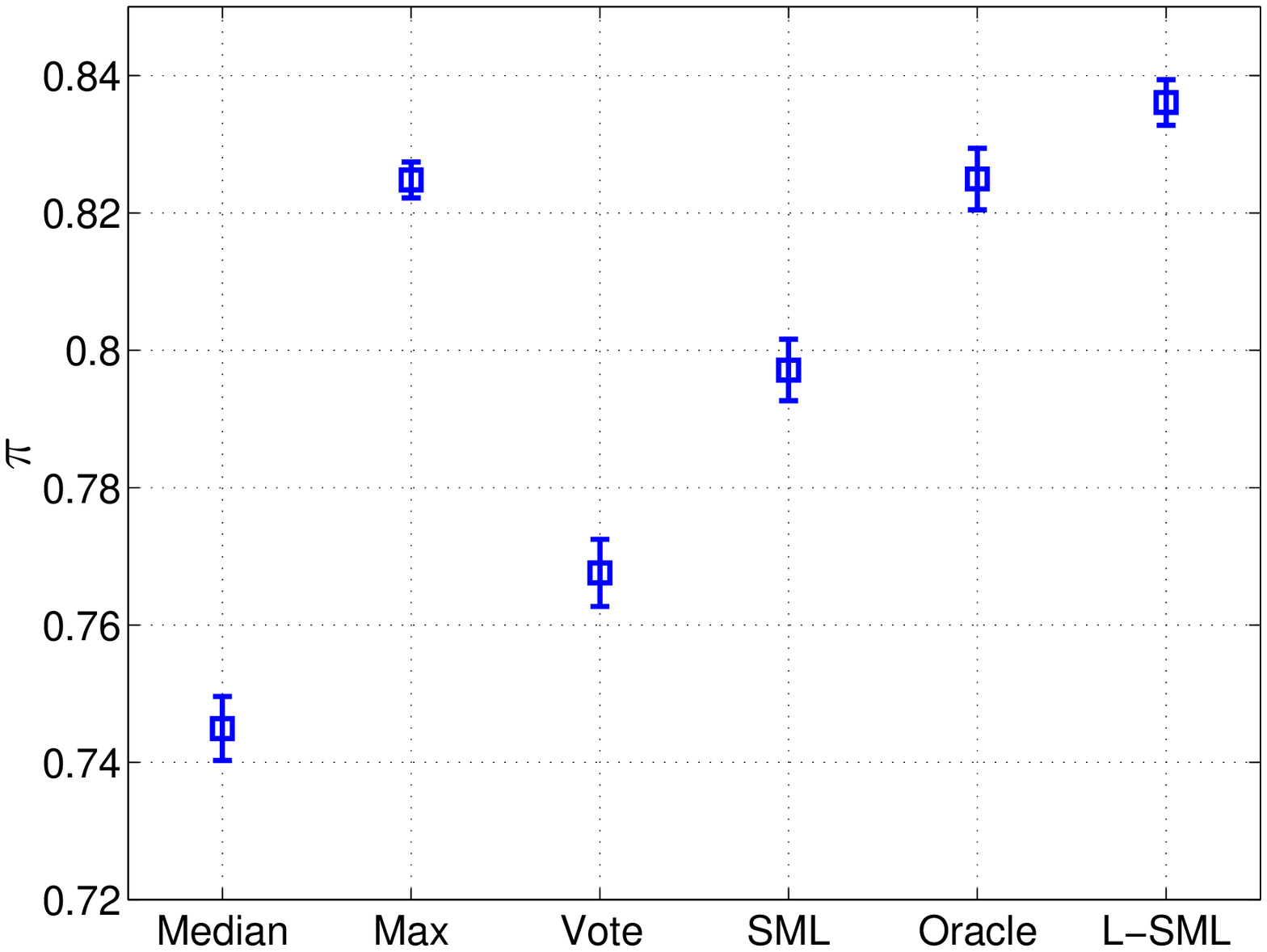}
                \caption{UCI magic dataset, a comparison of four unsupervised ensemble learners.}
                \label{Fig:UCI_magic}
        \end{minipage}
\end{figure}

To validate our theoretical analysis, we generated artificial 
binary data according to our assumed model, on a balanced classification problem with $b=0$. We generated an ensemble of $m=20$ classifiers with $n=10^4$ instances.
All the parameters of the ensemble were chosen uniformly at random from the following intervals:
$\Pr(\alpha=1|Y=1),\Pr(\alpha=-1|Y=-1) \in [0.5, 0.8], \{\psi_i^\alpha,\eta_i^\alpha\} \in [0.7, 0.9]$. 
We consider the case where there is only one group $G_1$ of correlated classifiers, with the remaining $m-|G_1|$ classifiers all conditionally independent. The size of the correlated group $|G_1|$ increases from $1$ to $10$.
Note that for $|G_1|=1$ all classifiers are conditionally independent.
Fig. \ref{Fig:Simulated} compares the balanced accuracy of the five unsupervised ensemble learners described above, as a function of the size of the first group, $|G_1|$.  As can be seen in Fig. \ref{Fig:Simulated}, up to $|G_1|=6$, the ensemble learner based on the concept of correlated classifiers achieves similar results to the optimal classifier ('oracle-L'). As expected from Lemma \ref{lem:score}, as $|G_1|$ increases, 
it is harder to correctly estimate $\mathbf c$ with the score matrix. 

A complementary graph which presents the probability to recover the correct assignment function as a function of $|G_1|$ appears in the appendix. As expected, the degradation in performance starts when the algorithm fails to correctly estimate the model structure.

\subsection{UCI data sets}
\label{subsec:UCI}
We applied our algorithms on various binary classification
problems using 4 datasets from the UCI repository:  Magic, Spambase, Miniboo and Musk. Our ensemble of $m = 16$ classifiers consists of 4 random forests, 3 logistic model trees, 4 SVM and 5 naive Bayes. Each classifier was trained on a separate, randomly chosen labeled dataset. In our unsupervised ensemble scenario we had access only to their predictions on a large independent test set.

 
We present results for the ’magic’ dataset, which contains $19000$ instances with $11$ attributes. The task is to classify each instance as
background or high energy gamma rays.
Further details and results on the other datasets appear in the appendix.


As seen in Fig. \ref{Fig:UCI_magic}, the L-SML improves substantially over the standard SML, and even on the oracle classifier that assumes conditional independence. Our method also outperforms the best individual classifier.

 In the appendix we show the conditional covariance matrix, Fig. \ref{Fig:magic_covariance},  and our assignment, Fig. \ref{Fig:magic_assign}. It can be observed that strongly dependent classifiers are indeed grouped together correctly.

\subsection{The DREAM mutation calling challenge}
\label{sub:dream}
The ICGC-TCGA DREAM challenge  is an international effort to improve standard methods for identifying cancer-associated mutations and rearrangements in whole-genome sequencing (WGS) data. This publicly available database contains both real and synthetic in-silico tumor instances. The database contains 14 different datasets, each with over 100,000 instances.

 Participants in the currently open competition are given access 
 to the predictions of about a hundred different classifiers (denoted there as pipe-lines)\footnote{The data can be downloaded from the challenge website http://dreamchallenges.org/}.
These classifiers were constructed by various labs worldwide, each employing their own biological knowledge and possibly proprietary labeled data. The two current challenges are to construct a meta-learner, by using either (1) all \(m\) classifiers; or (2) at most five of them. 
We evaluate the performance of the different meta-classifiers $f_{mc}$ by their balanced error,
\[
1-\pi = \tfrac{1}{2}(\Pr(f_{mc}=1|y=-1)+\Pr(f_{mc}=-1|y=1)).
\]
Below we present results on the datasets  S1, S2 and S3 for which ground-truth labels have been released.

\paragraph{Challenge \RNum{1}.}

The balanced errors of the different meta-learners, constructed using all $m$ classifiers, are given in table \ref{table:sub_ch_1}. 
 The L-SML method outperforms the other meta-learners in all the three datasets. On the S3 dataset, it reduces the balanced error by more than 20 \% over competing meta learners.

\paragraph{Challenge \RNum{2}}
Here the goal is to construct a sparse meta-learner based on at most five individual classifiers from the ensemble. 
For the methods based on the Dawid and Skene model (SML+EM, voting and Oracle-CI), we took the 5 classifiers with the highest estimated (or known) balanced accuracies.
For our model, since the estimated number of groups is larger than five, we first took the best classifier from each group, and then chose the five classifiers with highest estimated balanced accuracies. 
For all methods, the final prediction was made by a simple vote of the five chosen classifiers. Though potentially sub-optimal, we nonetheless chose it as our purpose was to compare the diversity of the different classifiers.

The results presented in table \ref{table:sub_ch_2} show that our method outperforms voting and SML, and are similar to those achieved by the oracle learner.

\begin{table}[t]                                 
        \centering                                        
        \begin{tabular}{|c|c|c||c|c|c|c|}                  
                \hline                                            
                & Mean & Best & Vote & \begin{tabular}{@{}c@{}}SML+ \\ EM\end{tabular}& \begin{tabular}{@{}c@{}}Oracle- \\ CI\end{tabular} & L-SML \\
                \hline                                            
                S1 & 6.1 & 1.7 & 2.8 & 1.7 & 1.7 & \bf{1.6} \\         
                \hline                                            
                S2 & 8.7 & 1.8 & 4.0 & 2.8 & 2.8 & \bf{2.3} \\         
                \hline                                            
                S3 & 8.3 & 2.5 & 4.3 & 2.3 & 2.3 & \bf{1.8} \\         
                \hline                                            
        \end{tabular}                                     
        \caption{Balanced error of meta-classifiers based on the full ensemble. For reference, the first two columns give the mean and smallest balanced error of all classifiers.}                          
        \label{table:sub_ch_1}                        
\end{table}           
\begin{table}[t]                    
        \centering                           
        \begin{tabular}{|c|c|c|c|c|}         
        \hline                               
        & Vote & SML+EM & Oracle-CI & L-SML \\
        \hline                               
        S1 & 3.2 & 2.3& \bf{1.9} &  2.0 \\        
        \hline                               
        S2 & 4.3 & 4.1 &\bf{2.5}  & 2.8 \\        
        \hline                               
        S3 & 2.9 & 2.9 & 2.8 & \bfseries{2.5} \\        
        \hline                                         
        \end{tabular}                        
        \caption{Balanced error of  sparse meta-classifiers. }             
        \label{table:sub_ch_2}           
\end{table}

\section{Acknowledgments}

This research was funded in part by the Intel Collaborative
Research Institute for Computational Intelligence
(ICRI-CI). Y.K. is supported by the National
Institutes of Health Grant R0-1 CA158167, and R0-1 GM086852. 



\newpage
\bibliographystyle{plain}
\bibliography{bib_unsupervised_classification}

\newpage

\onecolumn
\appendix

\section{Proof of Lemma \ref{lem:R}} 
This proof is based on the following lemma, which appears in \cite{Parisi_2014}:\\
If two classifiers $f_i,f_j$ are conditionally independent given the class label $Y$, then the covariance between them is equal to,
\begin{equation}\label{eq:r_2}
r_{ij} = (1-b^2)(\psi_i +\eta_i-1)(\psi_j +\eta_j-1).
\end{equation}
 In our model, if $\mathbf c(i) \neq \mathbf c(j)$, then $f_i,f_j$ are indeed conditionally independent (Fig. \ref{Fig:graph},right). The first part of lemma \ref{lem:R} follows directly from Eq. \eqref{eq:r_2}, with $v_i^{off}=\sqrt{1-b^2}(\psi_i +\eta_i-1)$.

To prove the second part of lemma  \ref{lem:R}, we note that according to our model, two classifiers $f_i,f_j$ with $\mathbf c(i)=\mathbf c(j)$ are conditionally independent given the value of their latent variable $\alpha$. Therefore, we can treat $\alpha$ as the class label, and apply Eq. \eqref{eq:r_2} with $b$ replaced by the expectation of $\alpha$, and the sensitivity and specificity $\psi_i, \eta_i$ replaced by $\psi_i^\alpha,\eta_i^\alpha$ respectively. 
Hence, Eq. \eqref{eq:r_2} becomes,
\begin{equation}
r_{ij} = (1-\mathbb E[\alpha]^2)(\psi_i^\alpha +\eta_i^\alpha-1)(\psi_j^\alpha +\eta_j^\alpha-1)=v_i^{on}v_j^{on},
\end{equation}
where $v_i^{on}=\sqrt{1-\mathbb E[\alpha]^2}(\psi_i^\alpha +\eta_i^\alpha-1)$.

\section{Proof of Lemma \ref{lem:sparsity}}
\label{Sec:lem_2}
We assume that $v^{on}$ and $v^{off}$ are sufficiently different in the following precise sense: We require  that for all 4 distinct indices $i,j,k,l$, $v^{on}_i\cdot v^{on}_j\cdot v^{on}_k\cdot v^{on}_l\neq v^{off}_i\cdot v^{off}_j\cdot v^{off}_k\cdot v^{off}_l$.

Next, we elaborate on the relation between $v_i^{off}$ and $v_i^{on}$.
Let us denote by $\psi_\alpha^y,\eta^y_\alpha$ the sensitivity and specificity of 
the latent variable $\alpha$. Let $f_i$ be a classifier that depends on $\alpha$. Applying Bayes rule, its overall sensitivity and specificity is given by,
\begin{eqnarray}
\psi_i  = \psi^y_\alpha \psi^\alpha_i +(1-\psi^y_\alpha)(1-\eta_i^\alpha) \notag\\
\eta_i  = \eta^y_\alpha \eta^\alpha_i +(1-\eta^y_\alpha)(1-\psi_i^\alpha).
\end{eqnarray}
Adding $\psi_i$ and $\eta_i$ we get the following,
\begin{equation}
\psi_i+\eta_i-1 = (\psi_\alpha^y+\eta_\alpha^y-1)(\psi_i^\alpha+\eta_i^\alpha-1).
\end{equation}
If $\mathbf c(i)=\mathbf c(j)$ we have the following dependency between $(v_i^{off},v_j^{off})$ and $(v_i^{on},v_j^{on})$,
\begin{equation}\label{eq:lin_dep}
\left[ \begin{array}{c}
v_i^{off}\\
v_j^{off}
\end{array}\right] = \sqrt{1-b^2}(\psi_\alpha^y+\eta_\alpha^y-1)
\left[
\begin{array}{c}
(\psi_i^\alpha+\eta_i^\alpha-1)\\
(\psi_j^\alpha+\eta_j^\alpha-1)
\end{array}\right] = \sqrt{\frac{1-b^2}{1-\mathbb E[\alpha]^2}}(\psi_\alpha^y+\eta_\alpha^y-1) 
\left[ \begin{array}{c}
v_i^{on}\\
v_j^{on}
\end{array}\right].
\end{equation} 
It follows that two elements $v_i^{off},v_j^{off}$ where $\mathbf c(i)=\mathbf c(j)$ are linearly dependent with the corresponding elements of $v_i^{on},v_j^{on}$. This fact shall be useful in proving the lemma.   

 To prove lemma \ref{lem:sparsity} we analyze all various possibilities for the group assignments of the four indices $i,j,k,l$ of 
\[M(i,j,k,l)=\det\left(
\begin{array}{cc}
	r_{ij} & r_{il} \\
	r_{kj} & r_{kl} 
\end{array}
\right)\].

\begin{enumerate}
        \item $c(i)=c(j)=c(k)=c(l)$: In this case $M(i,j,k,l)=v^{on}_iv^{on}_jv^{on}_kv^{on}_l-v^{on}_iv^{on}_lv^{on}_kv^{on}_j=0$.
        \item $c(i)\neq c(j)$, and $c(j)\neq c(k)$, and $c(k)\neq c(l)$ and $c(l)\neq c(i)$: Here $M(i,j,k,l)=v^{off}_iv^{off}_jv^{off}_kv^{off}_l-v^{off}_iv^{off}_lv^{off}_kv^{off}_j=0$.
        \item $c(i)=c(l)=c(k)\neq c(j)$: $M(i,j,k,l)=v^{off}_iv^{off}_jv^{on}_kv^{on}_l-v^{on}_iv^{on}_lv^{off}_kv^{off}_j=v^{off}_jv^{on}_l\left(v^{off}_iv^{on}_k-v_i^{on}v^{off}_k\right)$.  From the linear dependency shown in Eq. \eqref{eq:lin_dep}. $\left(v^{off}_iv^{on}_k-v_i^{on}v^{off}_k\right)=0$.
        \item $c(i)=c(j)$, $c(k)=c(l)$ and $c(i)\neq c(k)$: $M(i,j,k,l)=v^{on}_iv^{on}_jv^{on}_kv^{on}_l-v^{off}_iv^{off}_lv^{off}_kv^{off}_j\neq 0$ from our assumption.
\end{enumerate}
It can be seen that $M_{ijkl}$ is equal to zero \textit{only} if either three or more of the indices are equal (cases (1) and (2)) or all four pairs which appear in the determinant belong to different groups (case (3)).

\section{Algorithm for the ideal setting}
An immediate  conclusion from lemma \ref{lem:sparsity}, is that the indices $i,j,k$ and $l$ for which $M(i,j,k,l)=0$ depend only on the assignment function. This means we can compare the pattern of zeros for $M(i_1,j,k,l)$ and $M(i_2,j,k,l)$ to decide if $f_{i1}$ and $f_{i2}$ belong to the same group. If $c(i_1)=c(i_2)$ then $M(i_1,j,k,l)=0\iff M(i_2,j,k,l)=0$. On the other hand if $c(i_1)\neq c(i_2)$ and at least one of the indices $i_1$ and $i_2$ , w.l.o.g $i_1$, belongs to a group with more than one element, then we can find $j,k$ and $l$ such that $M(i_1,j,k,l)\neq0$ but $M(i_2,j,k,l)=0$. This occurs when $c(i_1)=c(j)$, and $c(i_2)\neq c(j)\neq c(k)\neq c(l)$.

This means that by comparing the pattern of zeros, we can recover the assignment function.  Notice, that according to the algorithm, all singleton classifiers, that is, classifiers who are conditionally independent with the rest of the ensemble, are grouped together under a common latent variable. This is not a problem, as our model is not unique and this is an equivalent probabilistic model, when the latent variable being identical to $Y$.

\begin{algorithm}[H]
	\caption{Check if $\mathbf c(i_1) = \mathbf c(i_2)$}
	\begin{algorithmic}[1]
		\State Initialize $(m-2) \times (m-3) \times (m-4)$ arrays $T_1,T_2$ to zero
		\For{$j \neq k \neq l \neq i_1,i_2$}
		\If{$r_{i_1j}r_{kl}-r_{i_1l}r_{kj}=0$} ($T_1(j,k,l)=1$) \EndIf
		\If{$r_{i_2j}r_{kl}-r_{i_2l}r_{kj}=0$} ($T_2(j,k,l)=1$) \EndIf
		\EndFor
		\If {($T_1=T_2$)} 
		\State $\mathbf c(i_1)=\mathbf c(i_2)$.
		\Else
		\State $\mathbf c(i_1)\neq\mathbf c(i_2)$.
		\EndIf
	\end{algorithmic}
	\label{algo_1}
\end{algorithm}

\section{Minimizing $\Delta$ is a NP hard problem}
We prove lemma \ref{lem:NP} for the case of $K=2$ clusters and known $\pmb v^{off},\pmb v^{on}$ vectors. Our goal is to find a minimizer for the following residual:
\begin{equation}
\hat {\mathbf c} = \argmin_{\mathbf c} \Delta(\mathbf c) = \argmin_{\mathbf c} \sum_{i,j} \mathds{1}_{\mathbf c}(i,j)(v_i^{on} v_j^{on}-r_{ij})^2 +(1-\mathds{1}_{\mathbf c}(i,j))(v_i^{off}v_j^{off} -r_{ij})^2 
\end{equation}
For the case of $K=2$ we can simplify the residual considerably. 
Let us define a vector $\pmb x \in \{-1,1\}^m$ where $x_i=1$ if $\mathbf c(i)=1$ and
$x_i=-1$ if $\mathbf c(i)=2$. 
We can replace the indicator function $\mathds{1}(i,j)$ with the following,
\begin{equation}
\mathds{1}(i,j) = \frac{(1+x_ix_j)}{2}, \qquad 1-\mathds{1}(i,j) = \frac{(1-x_ix_j)}{2}.
\end{equation}
In addition, we can replace the minimization over $\mathbf c$ with a minimization over $\pmb x$,
\begin{eqnarray}
\hat {\pmb x} &=& \argmin_{\pmb x} \sum_{i,j} \frac{(1+x_ix_j)}{2} (v_i^{on} v_j^{on}-r_{ij})^2 +\frac{(1-x_ix_j)}{2}(v_i^{off}v_j^{off} -r_{ij})^2 \notag \\
&=& \argmin_{\pmb x} \sum_{i,j} \frac{1}{2} \left((v_i^{on} v_j^{on}-r_{ij})^2+(v_i^{off}v_j^{off} -r_{ij})^2 \right) \notag \\
&& +\frac{x_ix_j}{2}\left((v_i^{on} v_j^{on}-r_{ij})^2+(v_i^{off}v_j^{off} -r_{ij})^2 \right).
\end{eqnarray}
The first term does not depend on $\pmb x$ and we can omit it from the minimization problem. Let us also define the matrix $\tilde R$,
\begin{equation}
\tilde r_{ij} = \frac{\left((v_i^{on} v_j^{on}-r_{ij})^2+(v_i^{off}v_j^{off} -r_{ij})^2 \right)}{2}
\end{equation}
We are left with the following minimization problem:
\begin{equation}\label{eq:np_hard_r}
\hat{\pmb x} = \argmin_{\pmb x} \sum_{i,j} x_ix_j \tilde r_{ij} = \argmin_{\pmb x}\pmb x' \tilde  R \pmb x
\end{equation}
If there is a binary vector whose residual is precisely zero, then it can be found by computing the eigenvector with smallest eigenvalue of the matrix $\tilde R$. If, however, the minimal residual is not zero, then eq. (\ref{eq:np_hard_r}) is a quadratic optimization problem involving discrete variables, which is well known to be a NP-hard problem.  

\section{Proof of Lemma \ref{lem:score}}

We start by proving the first part of the lemma, where $\mathbf c(i)=\mathbf c(j)$. The score matrix $s_{ij}$ is a sum of all possible $2 \times 2$ determinants,
\begin{equation}
s_{i,j} = \sum_{k,l \neq i,j} |r_{ij}r_{kl}-r_{il}r_{jk}| = \sum_{k,l \neq i,j} s_{ij}^{kl},
\end{equation}
where we define $s_{ij}^{kl}$ as a single score element.
The following table separates the group of $s_{ij}^{kl}$ score elements into three types, and states the number of elements in each type. 
\begin{table}[h!]
        \centering
        \begin{tabular}{cc}
                
                Element type & Number of elements \\
                $\mathbf c(i)= \mathbf c(j) \neq \mathbf c(k) \neq \mathbf c(l)$ & $m^2\left(1-\frac{3}{K}+\frac{2}{K^2}\right)$ \\
                $\mathbf c(i)= \mathbf c(j) = \mathbf c(k) \neq \mathbf c(l)$ & $m^2\left(\frac{1}{K}-\frac{2}{m}\right)\left(1-\frac{1}{m}\right)$ \\
                $\mathbf c(i)= \mathbf c(j) = \mathbf c(k) = \mathbf c(l)$ & $m^2\left(\frac{1}{K}-\frac{2}{m}\right)\left(\frac{1}{K}-\frac{3}{m}\right)$
        \end{tabular}
        
\end{table}

According to lemma \ref{lem:R}, the contribution to the score from elements of the second and third  type is exactly $0$ (see details in Sec. \ref{Sec:lem_2}). We will therefore focus on analyzing the score elements of the first type, where 
$\mathbf c(i)= \mathbf c(j) \neq \mathbf c(k) \neq \mathbf c(l)$. Recall, that we assume a symmetrical case where $b=0$, and $\Pr(\alpha=1|y=1)=\Pr(\alpha=-1|y=-1)$. These assumptions imply that  $\mathds{E}[\alpha_k]=0$ for all $k=1...K$.
Let us consider Lem. \ref{lem:R} in order to analyze the value of $s_{ij}^{kl}$, 
\begin{eqnarray}
s_{ij}^{kl} &=& |r_{ij}r_{kl}-r_{ij}r_{jk}| =  |(2\pi_i^\alpha-1)(2\pi_j^\alpha-1)(2\pi_k-1)(2\pi_l-1)-(2\pi_i-1)(2\pi_j-1)(2\pi_k-1)(2\pi_l-1)| \notag \\
 &=& |(2\pi_k-1)(2\pi_l-1)\big((2\pi_i^\alpha-1)(2\pi_j^\alpha-1)-(2\pi_i-1)(2\pi_j-1)\big)|
\label{eq:s_i_j_dev}
\end{eqnarray}
where $\pi_i^\alpha = \tfrac{1}{2}(\psi_i^\alpha+\eta_i^\alpha)$.
For simplicity of notation, let us denote by $\gamma$ the ratio of true positives and negatives of the latent variables:
\begin{equation}
\gamma = \Pr(\alpha_k=1|Y=1) = \Pr(\alpha_k=-1|Y=-1)
\end{equation}
It can easily be shown that the following holds:
\begin{equation}
(2\pi_i-1)= (2\gamma-1)(2\pi_i^\alpha-1) \qquad (2\pi_j-1)= (2\gamma-1)(2\pi_j^\alpha-1)
\label{eq:2_pi_1}
\end{equation}
Inserting \eqref{eq:2_pi_1} into \eqref{eq:s_i_j_dev} we get,
\begin{eqnarray}
s_{ij}^{kl} = |(2\pi_k-1)(2\pi_l-1)(2\pi_i^\alpha-1)(2\pi_j^\alpha-1)(1-(2\gamma-1)^2)| = \notag \\
|4(2\pi_k-1)(2\pi_l-1)(2\pi_i^\alpha-1)(2\pi_j^\alpha-1)(\gamma(1-\gamma))|
\label{ap:s_ij}
\end{eqnarray}

Let us now derive the values of the conditional covariance matrices $C^+,C^-$.
In order to obtain $C^+$, we can apply the first part of Lem.$\ref{lem:R}$, and replace the class imbalance $b$, which is the mean value of $Y$, with $\mathbb E[\alpha|Y=1]$. A similar argument applies to $C^-$. 
 The value for the conditional expectation of $\alpha$ is equal to,
\begin{equation}
\mathbb E[\alpha|Y=1] = 2\gamma-1 \quad \mathbb E[\alpha|Y=-1] = 1-2\gamma
\end{equation}
A simple derivation yields the following for both cases,
\begin{equation}
(1-\mathbb E[\alpha|Y=1]^2)=(1-\mathbb E[\alpha|Y=-1]^2)=4\gamma(1-\gamma)
\end{equation}
The value of $c^+_{ij}$ is therefor equal to $c^-_{ij}$, and both are equal to the following,
\begin{equation}
c^+_{ij}=c^-_{ij} = 4\gamma(1-\gamma)(2\pi_i^\alpha-1)(2\pi_j^\alpha-1)
\label{ap:C}
\end{equation}
Inserting \eqref{ap:C} into \eqref{ap:s_ij} we get the following,
\begin{equation}
s_{ij}^{kl} = |(2\pi_k-1)(2\pi_l-1)c^+_{ij}|=|(2\pi_k-1)(2\pi_l-1)c^-_{ij}|
\end{equation}
We will remain with $C^+$ for simplicity,
The total score contribution of the first type of elements is therefore,
\begin{equation}
 \sum_{k,l} s_{ij}^{kl} = |c^+_{ij}| \sum_{k,l}| (2\pi_k-1)(2\pi_l-1)|  
\end{equation}
Assuming $(2\pi_i-1)>\delta>0,\,\forall i$, the latter simplifies to,
\begin{equation}
s_{ij} > |c^+_{ij}| \delta^2 m^2(1-\tfrac{3}{K}+\tfrac{2}{K^2})>|c^+_{ij}| \delta^2 m^2(1-\tfrac{3}{K})
\end{equation}

We next turn to proving an upper bound when $\mathbf c(i)\neq\mathbf c(j)$. Once again we can separate the different elements into three types,
\begin{table}[h!]
\centering
        \begin{tabular}{cc}
                
                Element type & Number of elements \\
                $\mathbf c(i)\neq \mathbf c(j) \neq \mathbf c(k) \neq \mathbf c(l)$ & $m^2\left(1-\frac{5}{K}+\frac{6}{K^2}\right)$ \\
                $\mathbf c(i)\neq \mathbf c(j) = \mathbf c(k) \neq \mathbf c(l)$ & $2m^2\left(\frac{1}{K}-\frac{1}{m}\right)\left(1-\frac{2}{K}\right)$ \\
                $\mathbf c(i)\neq \mathbf c(j) = \mathbf c(k) = \mathbf c(l)$ & $m^2\left(\frac{1}{K}-\frac{2}{m}\right)\left(\frac{1}{K}-\frac{3}{m}\right)$
        \end{tabular}
\end{table}

The only contribution comes from the second type, as according to our model, if all indices come from different groups, or if three come from the same group, the determinant is equal to $0$ (see. \ref{Sec:lem_2}).
In addition, since $(2\pi_i-1)>\delta>0\, \forall i$, the values of $r_{ij}$ are positive for all $(i,j)$ pairs. Since $0<r_{ij}\leq 1$ for all score elements $s_{ij}^{kl}=|r_{ij}r_{kl}-r_{il}r_{kj}|\leq1$.
The total value of $s_{ij}$ is bounded by the following
\begin{equation}
s_{ij}\leq 2m^2\left(\frac{1}{K}-\frac{1}{m}\right)\left(1-\frac{2}{K}\right) < \frac{2m^2}{K}\left(1-\frac{2}{K}\right)
\end{equation}

\section{Additional results}

\subsection{Artificial data}
\begin{figure}[!t]      
        \begin{minipage}[b]{.45\textwidth}
                \centering      
                \includegraphics[width=\textwidth,height = 0.7\textwidth]{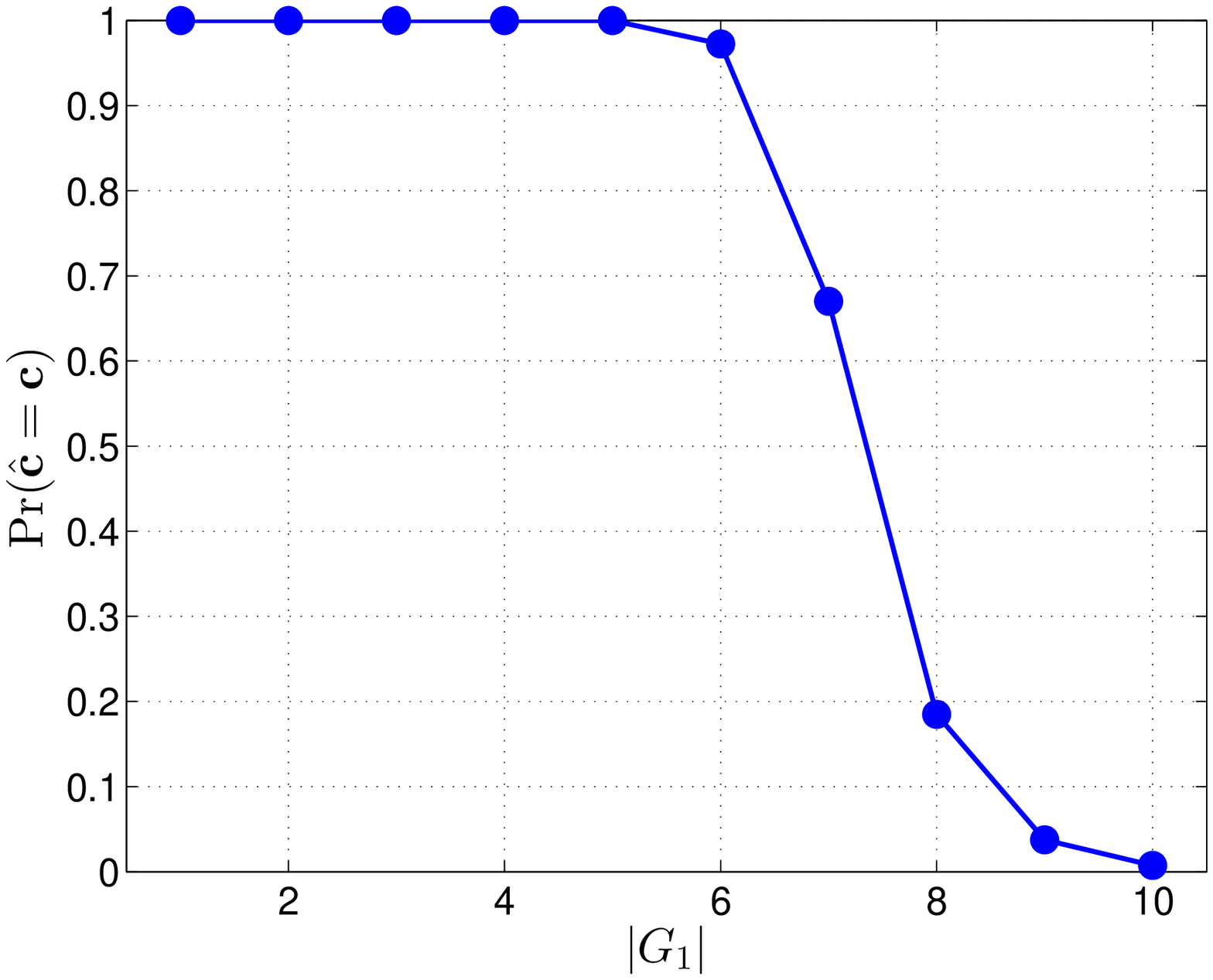}
                \caption{Probability of estimating the exact assignment function  $\mathbf c$ as a function of the size of the correlated group $|G_1|$}
                \label{Fig:percent}
        \end{minipage}
        \hspace{1cm}        
        \begin{minipage}[b]{.45\textwidth}
                \centering
                \includegraphics[width=\textwidth,height = 0.7\textwidth]{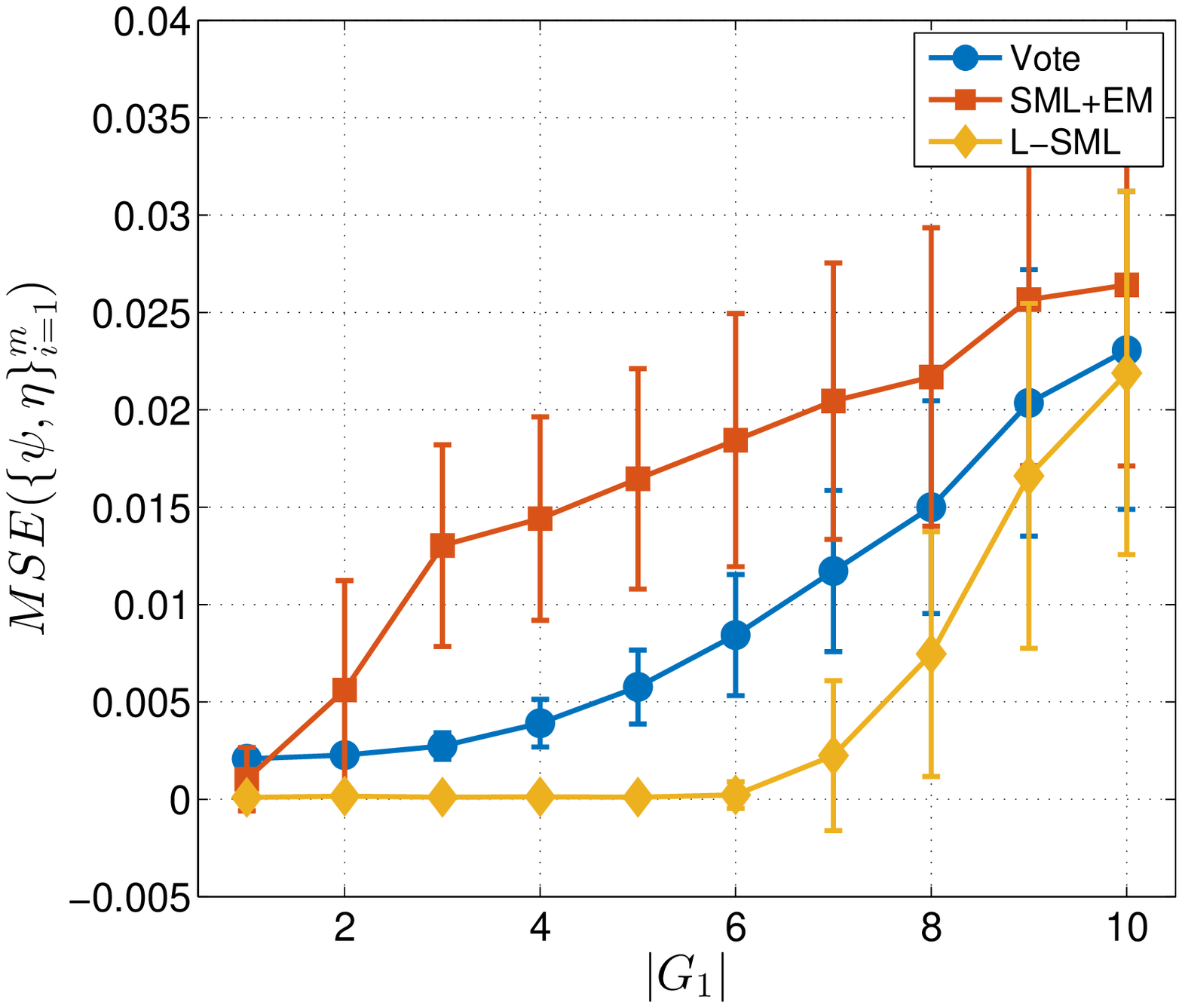}
               \caption{A comparison of the mean squared error in estimating the accuracies of the different classifiers in the ensemble.}
              \label{Fig:MSE_psi_eta}
        \end{minipage}
\end{figure}

In Fig. \ref{Fig:percent} we present  the probability of our spectral clustering based algorithm to recover both the correct number of classes $K$ and the correct assignment function $\mathbf c$, as a function of $|G_1|$. Up to $|G_1|=6$, our algorithm successfully estimates $\mathbf c$, with no errors. When $|G_1|>7$, the algorithm completely fails. The degradation in performance presented in Fig. \ref{Fig:Simulated}, corresponds to the point where the algorithm fails to estimate $\mathbf c$ correctly.

In Fig.  \ref{Fig:MSE_psi_eta} we present the mean squared error (MSE) of the sensitivity and specificity estimation  for the ensemble of classifiers,
as a function of $|G_1|$, defined as 
\begin{equation}
MSE(\{\psi,\eta\}_{i=1}^m)  = \tfrac{1}{2m} \sum_{i=1}^m \big(  (\hat \psi_{i} -\psi_{i})^{2}+(\hat \eta_{i}-\eta_{i})^2 \big). 
\end{equation} 
We compare the following three methods: (1) Majority vote ; (2) SML+EM; (3)L-SML.
It can be seen that the performance of the SML degrades very fast when the conditional independence assumption is violated. The performance of the L-SML is almost perfect up to the point where $|G_1|=6$, where as we have seen in Fig. \ref{Fig:percent}, the model is correctly estimated. The performance is still superior to other methods, even for large values of $|G_1|$. 

\subsection{UCI  results}
\begin{figure}[!t]      
        \begin{minipage}[b]{.45\textwidth}
                \centering                      
                \includegraphics[width=\textwidth]{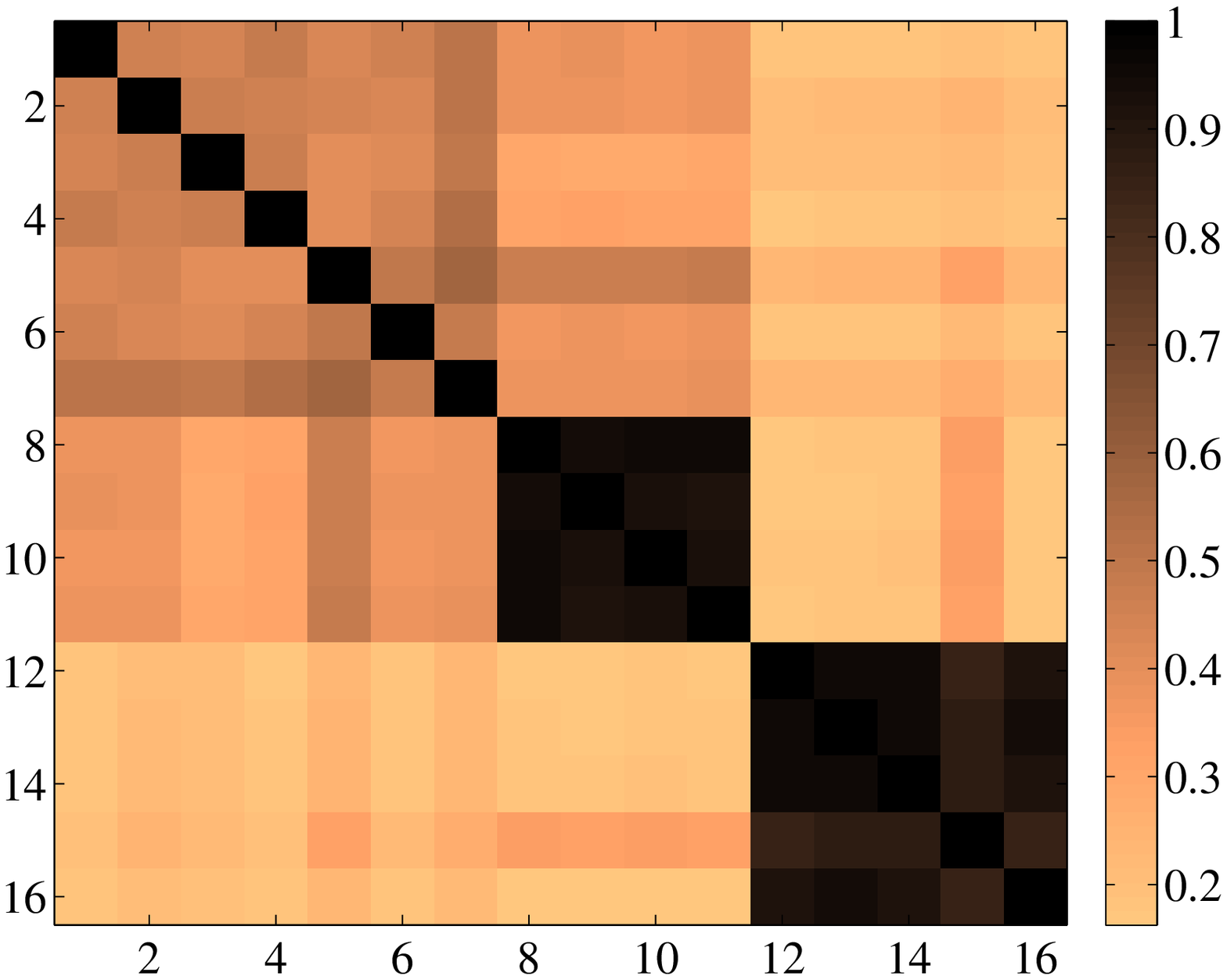}
                \caption{Magic database - conditional covariance  matrix $\frac{1}{2}(C^++C^-)$.}
                \label{Fig:magic_covariance}
        \end{minipage}
        \hspace{1cm}
        \begin{minipage}[b]{.45\textwidth}
                \centering
                \includegraphics[width=\textwidth]{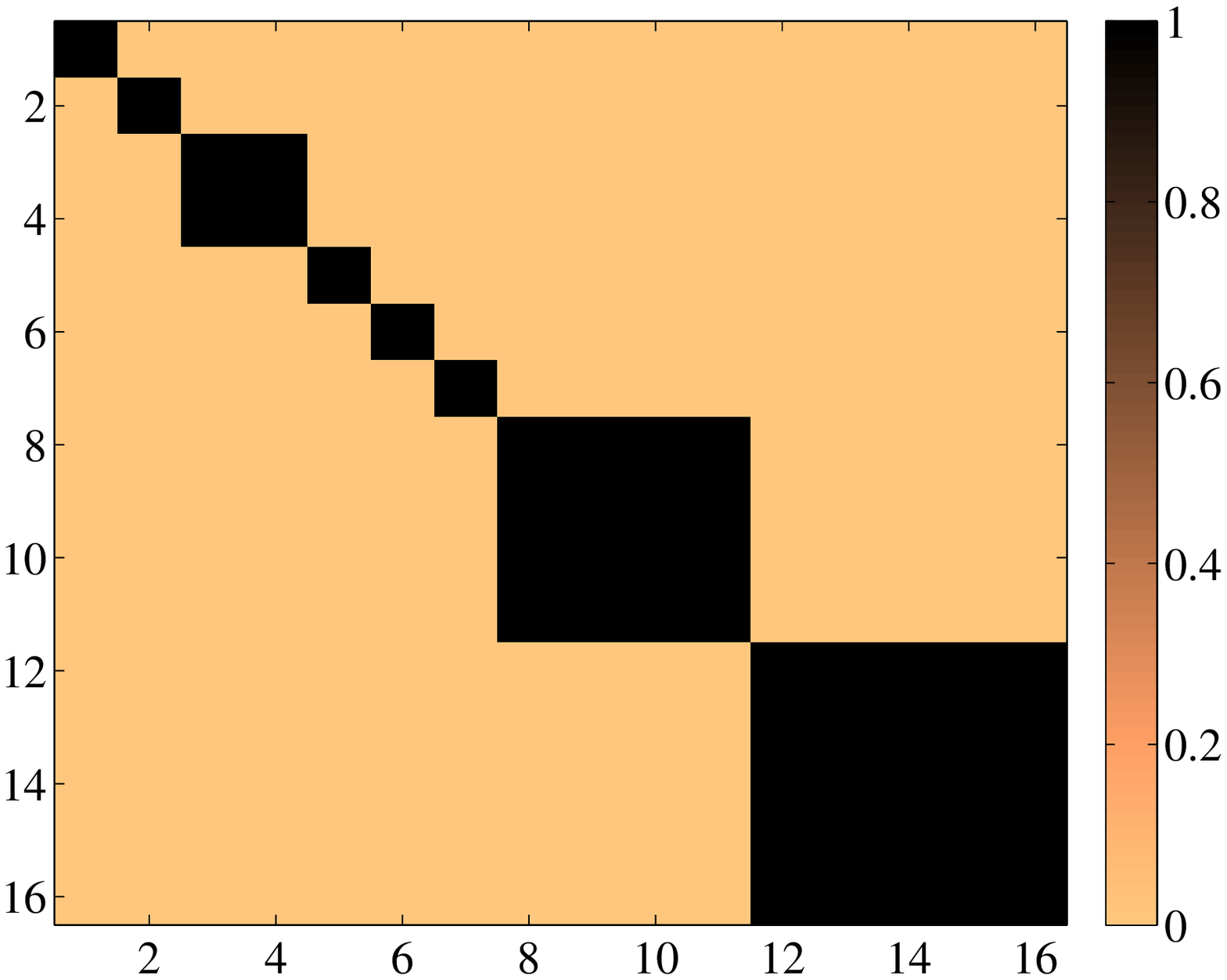}
                \caption{Magic database - The estimated group return by our algorithm.  }
                \label{Fig:magic_assign}
        \end{minipage}
\end{figure}

For the magic dataset, Fig. \ref{Fig:magic_covariance} presents the conditional covariance matrix $\tfrac{1}{2}(C^++C^-)$, which is unknown to us. The group of SVM classifiers (12-16) are highly dependent, as well as the group of naive Bayes classifiers (8-11). The groups of random forest classifiers and logistic model trees are weakly dependent. 

Fig.  \ref{Fig:magic_assign} presents an example of the estimated assignment function $\hat {\mathbf c}$ for the same dataset. The groups of SVM classifiers were assigned together, as well as the naive Bayes classifiers. Except for a single pair, the random forest and logistic model trees were assigned to separate groups.

\begin{figure}[!t]      
               \centering
        \begin{subfigure}[b]{.45\textwidth}
                \centering      
                \includegraphics[width=\textwidth]{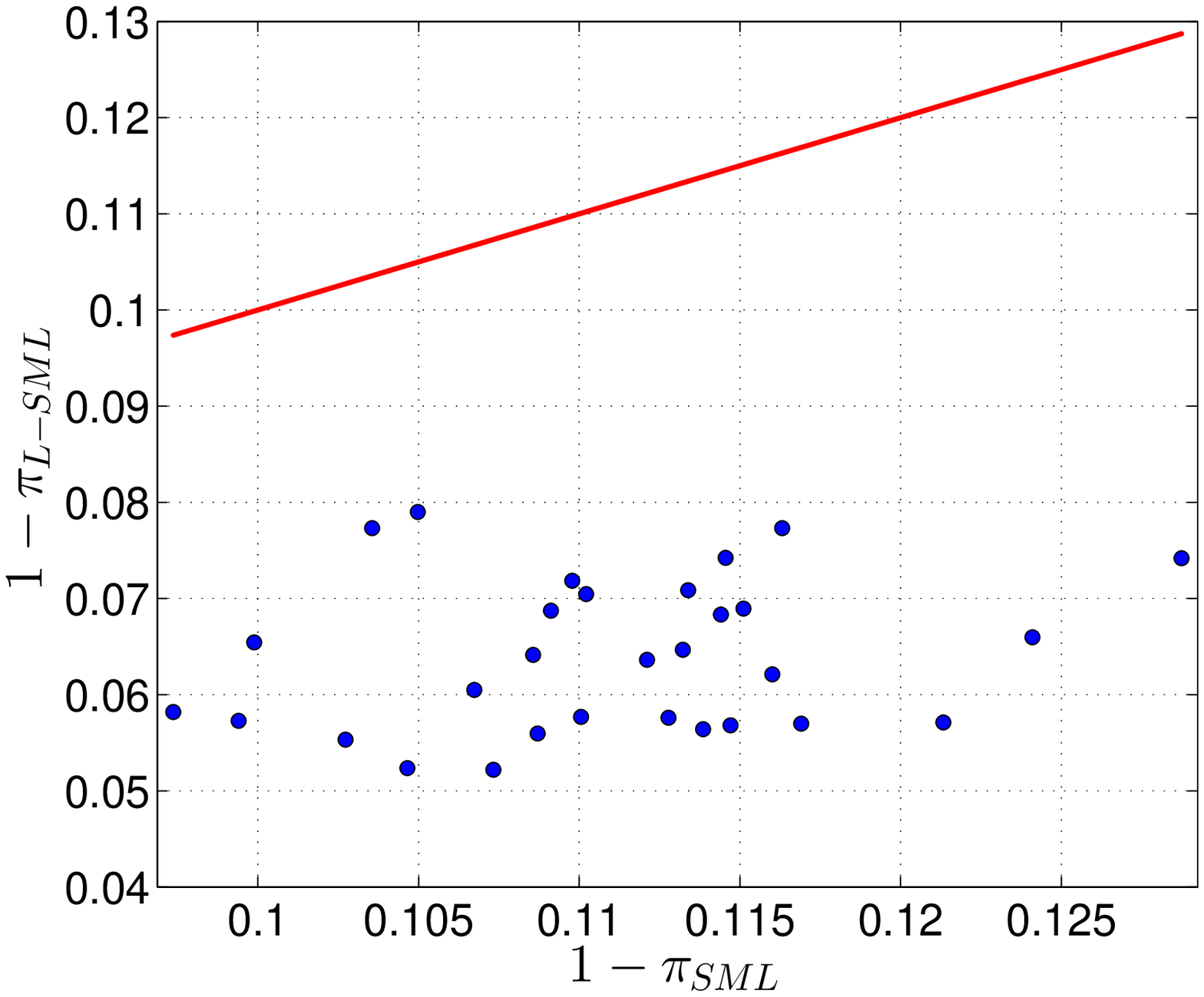}
                \caption{UCI Musk dataset, a comparison between the balanced error  of the SML and L-SML.}
                \label{Fig:musk}
        \end{subfigure}
        \hspace{1cm}
        \begin{subfigure}[b]{.45\textwidth}
                \centering
                \includegraphics[width=\textwidth]{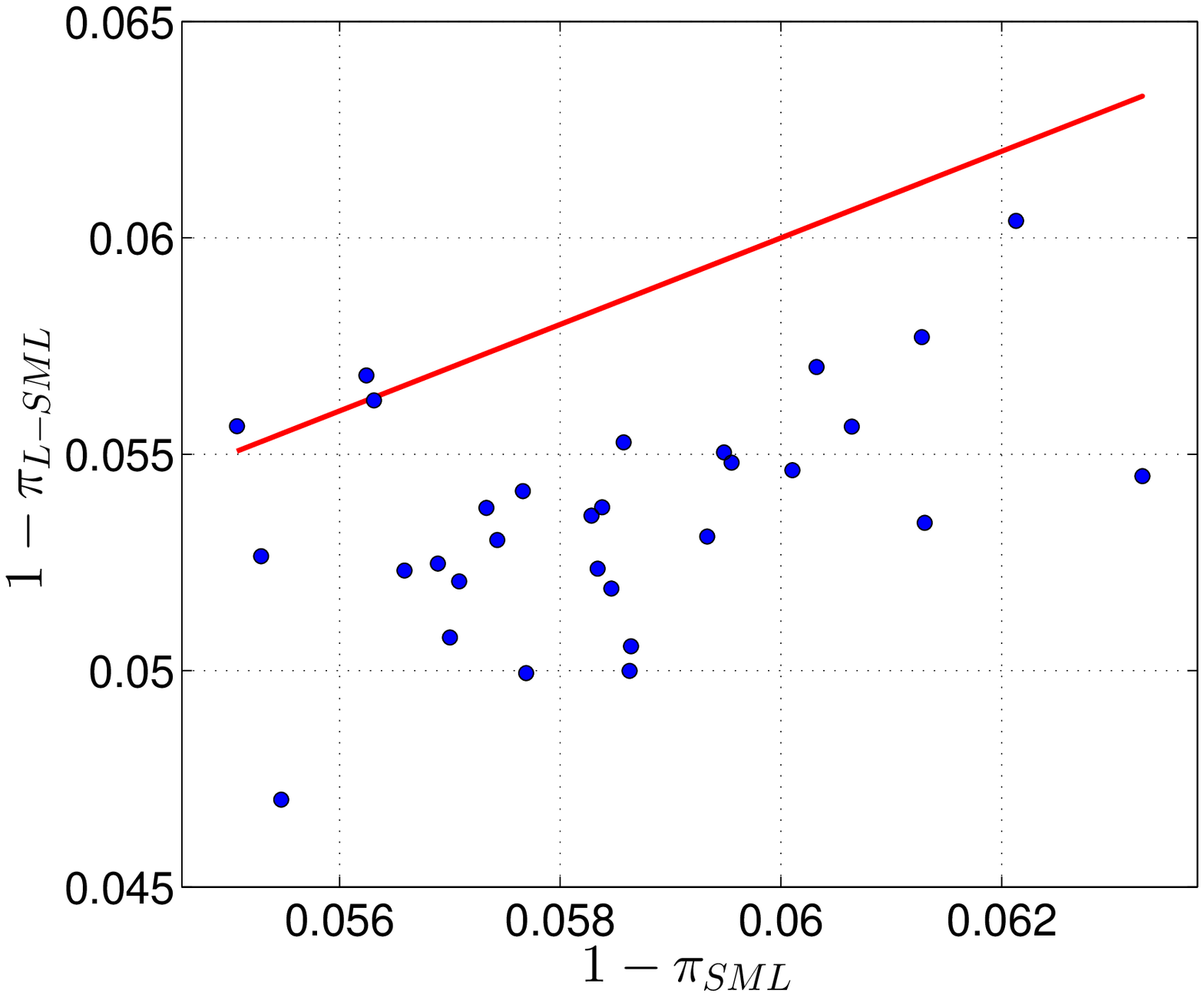}
                \caption{UCI Spambase dataset, a comparison between the balanced error  of the SML and L-SML.}
                \label{Fig:spam}
        \end{subfigure}
        \begin{subfigure}[b]{.45\textwidth}
        \centering
        \includegraphics[width=\textwidth]{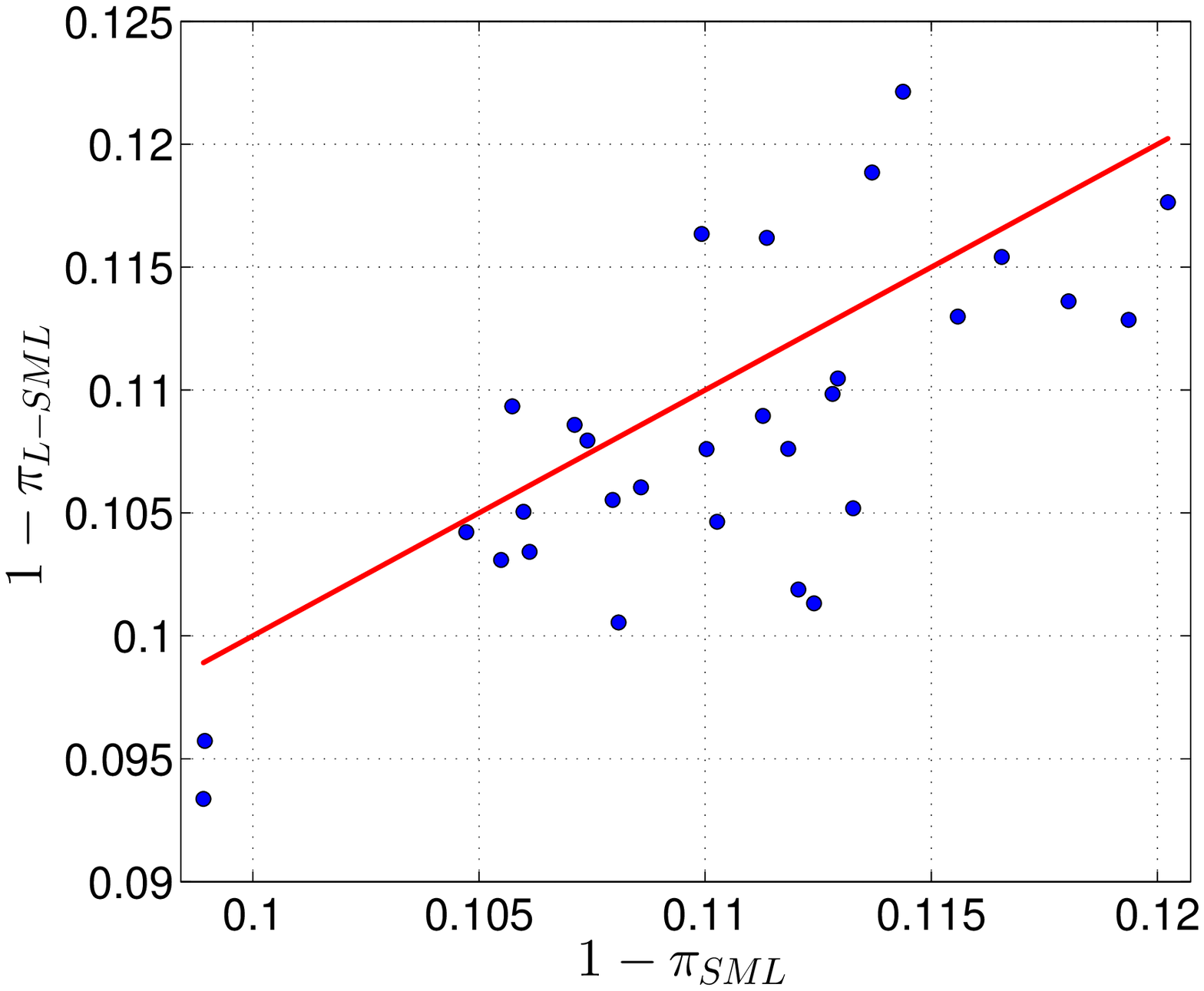}
        \caption{UCI Miniboo dataset, a comparison between the balanced error  of the SML and L-SML.\\}
        \label{Fig:miniboo}  
     \end{subfigure} 
     \hspace{1cm} 
     \begin{subfigure}[b]{.45\textwidth}
        \centering
        \includegraphics[width=\textwidth]{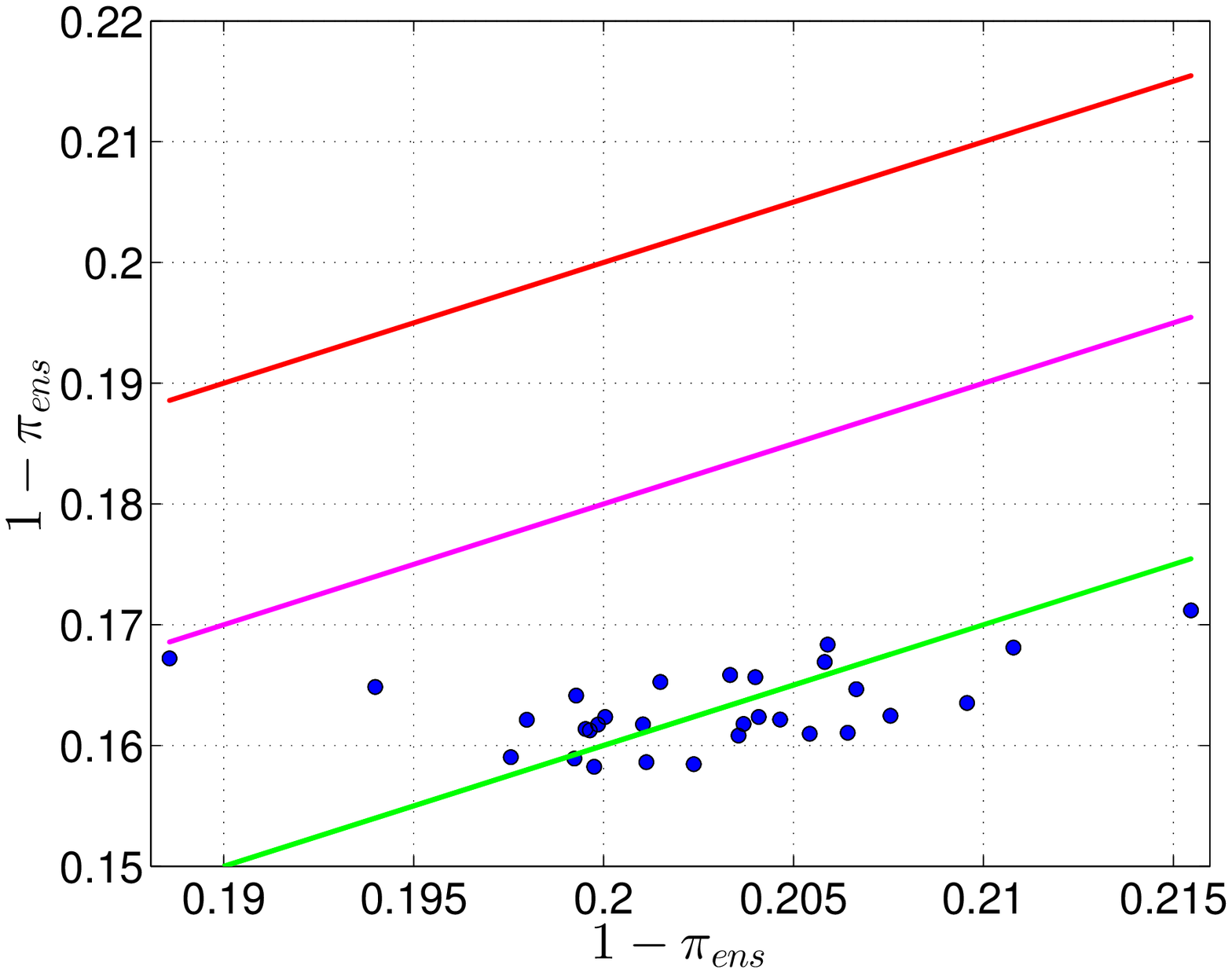}
        \caption{UCI Magic dataset. The magenta and green lines represent $2\%$ and $4\%$ balanced accuracy improvement over the SML results.}
        \label{Fig:magiccc}
     \end{subfigure}
             \caption{} 
\end{figure}

In figures \ref{Fig:musk},\ref{Fig:spam} and \ref{Fig:miniboo} we present the results for the following 3 additional datasets from the UCI repository:
\begin{itemize}
        \item Musk dataset - detection of certain types of molecules.
    \item Spam dataset - detection of spam from regular mail.
    \item Miniboo dataset - detection of electron neutrinos (signal) from muon neutrinos (background).
\end{itemize}

The base classifiers are identical to the ones used for the Magic dataset: (1) 4 random forest (2) 3 Logistic Model Trees (3) 4 SVM  (4) 5 naive Bayes .

In  figures \ref{Fig:musk}-\ref{Fig:magiccc}, the x-axis is the L-SML balanced error, and the y-axis is the SML balanced error. The results of multiple experiments, each time with the classifiers constructed using different random subset of labeled examples, are presented as blue dots, while the red line represents the $y=x$ line, i.e. when the error of the L-SML and SML are the same. For the Magic dataset, figure \ref{Fig:magiccc}, we add two lines which represent $2\%$ and $4\%$ improvement over the standard SML.

We can see in the figures that the improvement due to explicit modeling of possible classifier dependencies is consistent across all datasets. The amount of improvement changes, however from dataset to dataset. The following table presents a summary of the different properties of the datasets together with the average improvement in the balanced accuracy between the two methods.

\begin{table}[!h]
        \centering
        \begin{tabular}{cccc}
                
                Dataset & Number of instances & number of features  & Mean difference\\
                Magic & $19000$ & $11$ & $4\%$ \\
                Spam & $4600$ & $57$ & $0.5\%$ \\
                Miniboo & $130000$ & $50$ & $0.2\%$ \\
                Musk & $6600$ & $168$ & $4.7\%$ \\
                
\end{tabular}
\label{Tab:UCI}
\end{table}

\end{document}